\documentclass[sigplan,twocolumn,10pt]{acmart}
\acmSubmissionID{4}
\usepackage{graphicx}
\usepackage{subfigure} 
\usepackage{subcaption}

\usepackage{pifont}
\newcommand{\cmark}{\ding{51}}
\newcommand{\xmark}{\ding{55}}

\usepackage{makecell}
\usepackage{multirow}

\usepackage{listings}

\usepackage{soul}
\usepackage{color}
\usepackage{xcolor}

\soulregister{\cite}7 % 注册\cite命令
\soulregister{\citep}7 % 注册\citep命令
\soulregister{\citet}7 % 注册\citet命令
\soulregister{\ref}7 % 注册\ref命令
\soulregister{\pageref}7 % 注册\pageref命令

\definecolor{commentColor}{RGB}{53,129,34}
\definecolor{keywordColor}{RGB}{172, 62, 158}
\definecolor{stringColor}{RGB}{194, 62, 42}
\definecolor{preprocessorColor}{RGB}{114, 75, 48}
\definecolor{characterColor}{RGB}{31, 53, 207}
\definecolor{numberColor}{RGB}{166, 166, 166}
\definecolor{oglobalColor}{RGB}{97, 64, 154}
\definecolor{globalColor}{RGB}{89, 127, 134}
\definecolor{functionColor}{RGB}{56,36,124}

\usepackage{xcolor}

\usepackage[ruled,linesnumbered]{algorithm2e} % http://www.ctan.org/pkg/algorithm2e
\usepackage{xcolor} % http://www.ctan.org/tex-archive/macros/latex/contrib/xcolor
\usepackage{tikz}
\usetikzlibrary{fit,calc}
\newcommand*{\tikzmk}[1]{\tikz[remember picture,overlay,] \node (#1) {};\ignorespaces}
\newcommand{\boxit}[5]{\tikz[remember picture,overlay]{\node[yshift=3pt,fill=#1,opacity=.25,fit={($(A)+(#2\linewidth,#3\baselineskip)$)($(B)+(#4\linewidth,#5\baselineskip)$)}] {};}\ignorespaces}
\newcommand{\subboxit}[5]{\tikz[remember picture,overlay]{\node[yshift=3pt,fill=#1,opacity=.25,fit={($(AA)+(#2\linewidth,#3\baselineskip)$)($(BB)+(#4\linewidth,#5\baselineskip)$)}] {};}\ignorespaces}
\colorlet{mypink}{red!40}
\colorlet{myblue}{cyan!40}
\colorlet{mygreen}{green!50}
\colorlet{myyellow}{yellow!60}
\colorlet{mygrey}{gray!30}

\AtBeginDocument{%
  }

\acmYear{2025}\copyrightyear{2025}
\setcopyright{acmlicensed}
\acmConference[EuroSys '25]{Twentieth European Conference on Computer Systems}{March 30--April 3, 2025}{Rotterdam, Netherlands}
\acmBooktitle{Twentieth European Conference on Computer Systems (EuroSys '25), March 30--April 3, 2025, Rotterdam, Netherlands}
\acmDOI{10.1145/3689031.3717455}
\acmISBN{979-8-4007-1196-1/25/03}

\begin{document}

%%
%% The "title" command has an optional parameter,
%% allowing the author to define a "short title" to be used in page headers.
\title{Samoyeds: Accelerating MoE Models with Structured Sparsity Leveraging Sparse Tensor Cores}
% \title{Samoyeds: \underline{A}ccelerating \underline{MoE} Models with \underline{S}tructure\underline{d} \underline{S}parsit\underline{y} Leveraging Sparse Tensor Cores}
%%
%% The "author" command and its associated commands are used to define
%% the authors and their affiliations.
%% Of note is the shared affiliation of the first two authors, and the
%% "authornote" and "authornotemark" commands
%% used to denote shared contribution to the research.
% \author{Ben Trovato}
% \authornote{Both authors contributed equally to this research.}
% \email{trovato@corporation.com}
% \orcid{1234-5678-9012}

\author{Chenpeng Wu}
% \authornotemark[1]
\authornote{Equal Contribution}
\email{cpwu_sjtu@sjtu.edu.cn}
\affiliation{%
  \institution{Shanghai Jiao Tong University}
  \state{Shanghai}
  \country{China}
}

\author{Qiqi Gu}
\authornotemark[1]
\email{qiqi.gu@sjtu.edu.cn}
\affiliation{%
  \institution{Shanghai Jiao Tong University}
  \state{Shanghai}
  \country{China}
}

\author{Heng Shi}
\email{heng.shi@sjtu.edu.cn}
\authornotemark[2]
\affiliation{%
  \institution{Shanghai Enflame Technology Co.Ltd; Shanghai Jiao Tong University}
  \state{Shanghai}
  \country{China}
}

\author{Jianguo Yao}
\email{jianguo.yao@sjtu.edu.cn}
% \authornotemark[2]
\authornote{Corresponding author}
\affiliation{%
  \institution{Shanghai Jiao Tong University}
  \state{Shanghai}
  \country{China}
}

\author{Haibing Guan}
\email{hbguan@sjtu.edu.cn}
\affiliation{%
  \institution{Shanghai Jiao Tong University}
  \state{Shanghai}
  \country{China}
}

% \authornotetext[1]{These authors contributed equally to this work.}

%%
%% By default, the full list of authors will be used in the page
%% headers. Often, this list is too long, and will overlap
%% other information printed in the page headers. This command allows
%% the author to define a more concise list
%% of authors' names for this purpose.
% \renewcommand{\shortauthors}{Wu et al.}
\renewcommand{\shortauthors}{Chenpeng Wu, Qiqi Gu, Heng Shi, Jianguo Yao, Haibin Guan.}
\renewcommand{\shorttitle}{Samoyeds: Accelerating MoE Models with Structured Sparsity Leveraging SpTCs}

\begin{abstract}
    The escalating size of Mixture-of-Experts (MoE) based Large Language Models (LLMs) presents significant computational and memory challenges, necessitating innovative solutions to enhance efficiency without compromising model accuracy. Structured sparsity emerges as a compelling strategy to address these challenges by leveraging the emerging sparse computing hardware. Prior works mainly focus on the sparsity in model parameters, neglecting the inherent sparse patterns in activations. This oversight can lead to additional computational costs associated with activations, potentially resulting in suboptimal performance.

    This paper presents Samoyeds, an innovative acceleration system for MoE LLMs utilizing Sparse Tensor Cores (SpTCs). Samoyeds is the first to apply sparsity simultaneously to both activations and model parameters. It introduces a bespoke sparse data format tailored for MoE computation and develops a specialized sparse-sparse matrix multiplication kernel. 
    Furthermore, Samoyeds incorporates systematic optimizations specifically designed for the execution of dual-side structured sparse MoE LLMs on SpTCs, further enhancing system performance.
    Evaluations show that Samoyeds outperforms SOTA works by up to 1.99$\times$ at the kernel level and 1.58$\times$ at the model level.
    Moreover, it enhances memory efficiency, increasing maximum supported batch sizes by 4.41$\times$ on average. 
    Additionally, Samoyeds surpasses existing SOTA structured sparse solutions in both model accuracy and hardware portability.
\end{abstract}

%%
%% The code below is generated by the tool at http://dl.acm.org/ccs.cfm.
%% Please copy and paste the code instead of the example below.
%%


%%
%% Keywords. The author(s) should pick words that accurately describe
%% the work being presented. Separate the keywords with commas.
%% A "teaser" image appears between the author and affiliation
%% information and the body of the document, and typically spans the
%% page.
% \begin{teaserfigure}
%   \includegraphics[width=\textwidth]{sampleteaser}
%   \caption{Seattle Mariners at Spring Training, 2010.}
%   \Description{Enjoying the baseball game from the third-base
%   seats. Ichiro Suzuki preparing to bat.}
%   \label{fig:teaser}
% \end{teaserfigure}

% \received{20 February 2007}
% \received[revised]{12 March 2009}
% \received[accepted]{5 June 2009}

\maketitle

\section{Introduction}

The emergence of ChatGPT has marked a pivotal milestone in the development of Large Language Models (LLMs), positioning them as a leading approach in deep learning. As LLMs evolve, there has been a notable escalation in both the scale of model parameters and computational demands\cite{brown2020language, dosovitskiy2020image, raffel2020exploring}. Recent models, such as LLaMA3\cite{touvron2023llama}, feature an impressive 400 billion parameters, a substantial increase from earlier models like BERT-base\cite{devlin2018bert}, which had only 110 million parameters. This rapid evolution presents significant challenges for deploying these LLMs within existing AI infrastructure, particularly given the limitations imposed by the pace of hardware advancements.

Moreover, the adoption of novel architectures, particularly the Mixture-of-Experts (MoE) technique\cite{GShard}, introduces additional complexity. The MoE layer, which consists of multiple experts, has been widely integrated into emerging LLMs due to its ability to enhance generalization\cite{GShard, Switch_Transformers, jiang2024mixtral} and manage multi-modal tasks effectively\cite{mustafa2022multimodal, team2023gemini}. This architectural innovation introduces unique demands for storage, bandwidth, and computation resources. Addressing the increasing scale and the novel architecture of LLMs is critical for their efficient deployment on contemporary AI accelerators. This underscores the need for developing evolutionary systems and methodologies to alleviate these pressures while fully utilizing existing hardware.

To address these challenges, sparse computing\cite{CunDS89, HassibiSW93} has emerged as a promising methodology to reduce memory footprint and computational costs by eliminating zero or least important elements. However, general sparsity, or unstructured sparsity\cite{cusparse, Sputnik}, is inefficient to implement on modern AI accelerators, particularly GPUs. The core issue with unstructured sparsity lies in the irregular pattern of non-zero elements, which complicates scheduling and balancing in SIMD programming models. This irregularity significantly hinders the utilization of performance-critical features in modern GPUs, such as coalesced memory access and warp-level synchronization. To overcome these limitations, structured sparsity has been introduced\cite{nmSparse, mao2017exploring, narang2017exploring}, effectively eliminating performance issues by representing data with regular sparse patterns. Also, this new paradigm of sparse computing has been supported by NVIDIA GPUs in its Sparse Tensor Cores (SpTCs) since the Ampere architecture, featuring a 2$\times$ peak performance boost compared to its dense counterparts\cite{mishra2021accelerating}. This support makes structured sparsity a viable hardware-software co-design solution for tackling the performance challenges of LLMs.

Currently, structured sparsity is primarily utilized to encode deep learning model parameters, significantly reducing model storage and computational workloads. This approach is exemplified by methods like cuSPARSELt\cite{cusparselt} and VENOM\cite{VENOM}. However, this design overlooks a critical aspect of modern LLMs: the inherent sparse pattern in activations, primarily induced by the routing mechanism of MoE layers. Neglecting these features represents a missed opportunity for further optimizations in LLM applications. It is noteworthy that the underlying sparse pattern in activations is well-recognized and investigated in state-of-the-art (SOTA) works\cite{megablocks,vllm}. However, none of these works effectively integrates this design principle with the capabilities of hardware.

To address these problems, we propose the Samoyeds system, which leverages sparsity in model parameters and activations simultaneously, without introducing extra overhead. 
To be specific, Samoyeds introduces a novel dual-side sparse data format, where one side represents the structured sparsity in model parameters, and the other side captures the dynamic sparsity that emerges during token routing in the MoE computation. 
By respecting the unique instruction requirements of SpTC and memory access pattern in dual-side structured sparse computation in MoE model, a specialized kernel execution scheme is proposed, which incorporates a series of systematic optimizations, including tiling orchestration, data stationary management, data packing reorganization, and layout optimization.

In summary, this paper makes the following contributions:
\begin{itemize}
    \item We revisit the underlying sparse patterns in large-scale MoE LLMs and explore the potential optimization by harnessing the dual-side sparse pattern untouched by existing SOTA works. This insight identifies a hardware-software co-optimization opportunity by leveraging the structured sparse computing facilities supported by NVIDIA Sparse Tensor Cores.

    \item We introduce a novel sparse data format for MoE computation with more flexible sparsity configuration and more efficient memory access for the dual-side sparse pattern. A corresponding execution scheme of sparse-sparse matrix multiplication kernels is proposed to improve hardware utilization.

    \item We implement several optimizations specifically tailored for this sparse format within the MoE execution, including hierarchical tiling, data stationary improvement, and data packing reorganization, collectively ensuring optimal computational efficiency.

    \item We extend Samoyeds kernel with various data layout configuration, meeting different requirements of matrix multiplication operands inside the MoE module, allowing seamless integration with existing MoE LLMs, and minimizing the overall memory I/O overhead.

    \item Experimental results show that the Samoyeds kernel outperforms SOTA sparse libraries, achieving up to 18.76$\times$ speedup over Sputnik (unstructured) and 1.99$\times$ over VENOM (structured). At the model level, Samoyeds surpasses the SOTA vLLM framework by up to 1.58$\times$ while increasing the average maximum batch size by 4.41$\times$. 
\end{itemize}

\section{Background}

\subsection{Mixture-of-Experts (MoE)}

\begin{figure}[ht]
  \centering
  \includegraphics[width=0.95\linewidth]{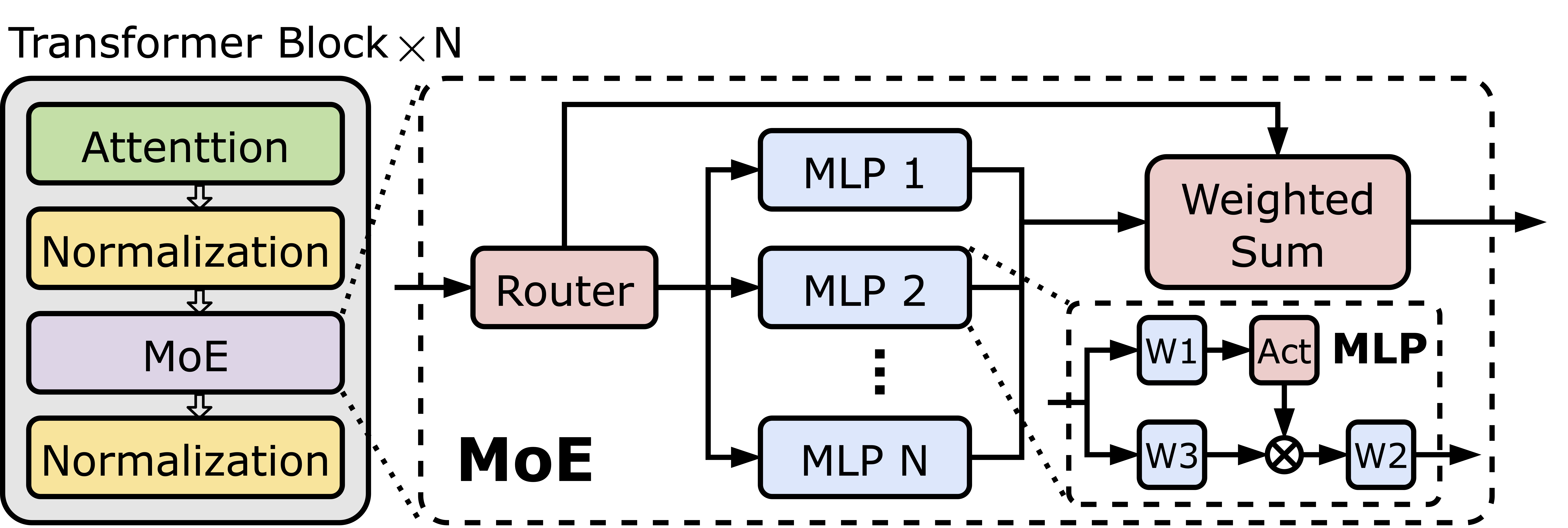}
  \caption{MoE LLM Architecture.}
  \label{fig:MoE结构}
\end{figure}

\begin{figure}[t]
  \centering
  \includegraphics[width=0.95\linewidth]{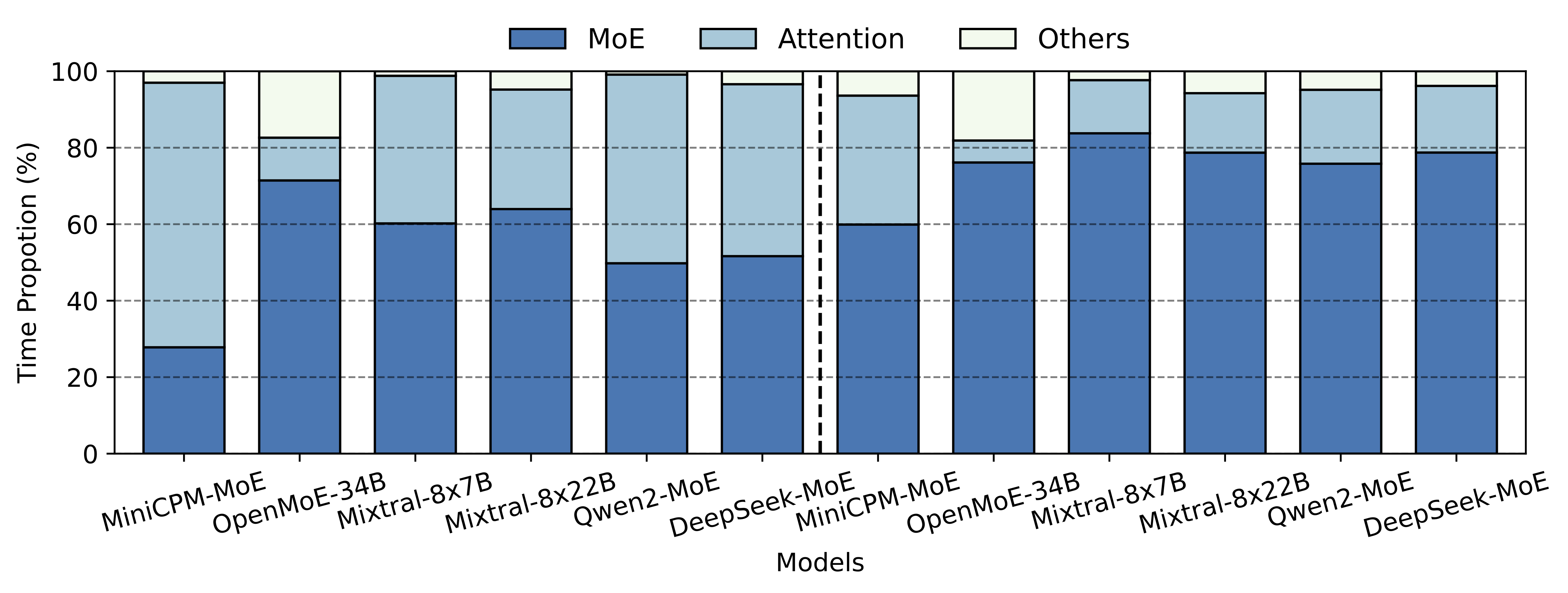}
  \caption{Time Breakdown of MoE Models. \textnormal{Left: Without Flash-Attention; Right: With Flash-Attention.}}
  \label{fig:MoE_Portion}
\end{figure}

\begin{figure}[t]
  \centering
  \includegraphics[width=0.95\linewidth]{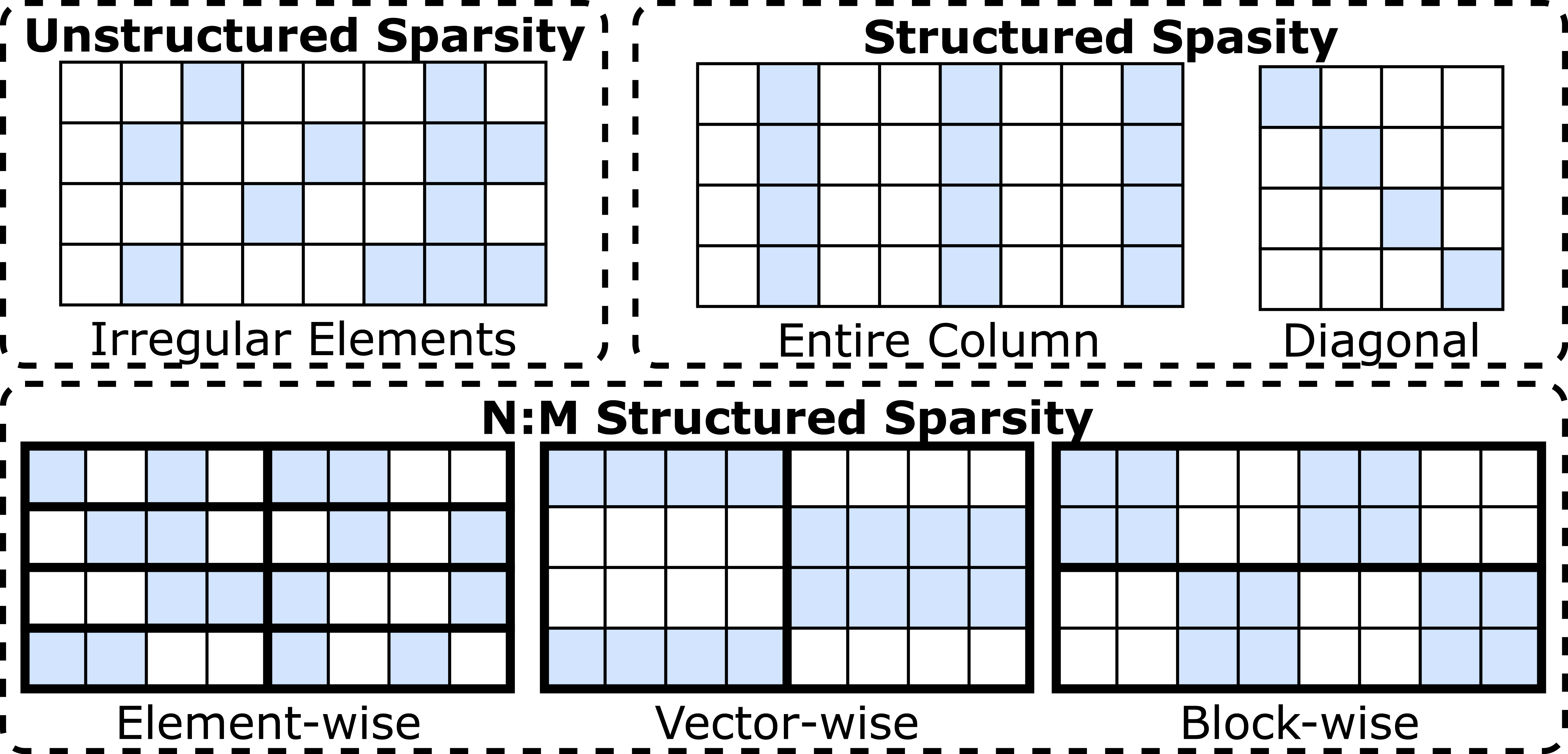}
  \caption{Data Patterns in Different Sparse Formats. \textnormal{Blank cells represent sparse elements.}}
  \label{fig:稀疏介绍}
\end{figure}

The typical MoE-based models, as shown in Figure \ref{fig:MoE结构}, consist of multiple Transformer Blocks\cite{GShard, Switch_Transformers, sanseviero2023moe, xue2024openmoe}. Each block includes attention, MoE and normalization layers. Within the MoE layer, a routing mechanism selects appropriate experts for each token. Tokens are then processed by Multilayer Perceptron (MLP) layers with multiple linear projections, called experts. The outputs of these experts are then propagated to the final output through a weighted sum. 

Unlike traditional models that compute across all activations, MoE models enhance efficiency by selectively activating experts, enabling the computation of partial activations to be skipped\cite{ZhouLLDHZDCLL22}. The MoE architecture is promising as it improves model generality without increasing demands on storage and computation resources\cite{ShazeerMMDLHD17, GShard, Switch_Transformers}. Research demonstrates that MoE models achieve competitive performance to larger dense models\cite{DeepSpeed_MoE}.

Numerous studies, such as Flash-Attention\cite{dao2022flashattention, dao2023flashattention2} and KV cache\cite{pope2023efficiently}, have optimized the attention layer, however, the MoE layer has received less focus. The execution time breakdown for a transformer block is illustrated in Figure \ref{fig:MoE_Portion}. Notably, the MoE layer, in most models, accounts for over half of the total processing time. With Flash-Attention enabled, the proportion of time consumed by the MoE layer exceeds 80\% in these models. This phenomenon underscores the urgent need for optimizations of the MoE layer to enhance overall model efficiency.

\subsection{Sparse Data Formats} \label{sec:sparse_data_formats}

Sparse computation employs different formats for data representation, providing a more efficient alternative to dense formats. These formats can be broadly categorized into two types: unstructured and structured, as illustrated in Figure \ref{fig:稀疏介绍}. Unstructured formats such as Coordinate List (COO) and Compressed Sparse Row (CSR) organize non-zero elements without constraints of regular patterns. However, this flexibility complicates GPU processing as it impedes efficient parallel execution. Structured formats, which organize data into regular patterns like columns or diagonals, allow for efficient GPU processing but potentially lose critical features\cite{Sputnik, mao2017exploring, narang2017exploring}.

N:M structured sparsity\cite{nmSparse, sun2021dominosearch, holmes2021nxmtransformer} format imposes additional constraints on structured formats by requiring the retention of N units within every contiguous set of M units. These units can vary in granularity, ranging from individual elements to vectors or blocks, as illustrated in Figure \ref{fig:稀疏介绍}. This format offers predictable patterns that facilitate efficient hardware implementation on accelerators, without compromising the flexibility and capability of feature representation. Given these advantages, our discussion will primarily focus on the N:M structured sparsity format.

\begin{figure}[t]
  \centering
  \includegraphics[width=0.98\linewidth]{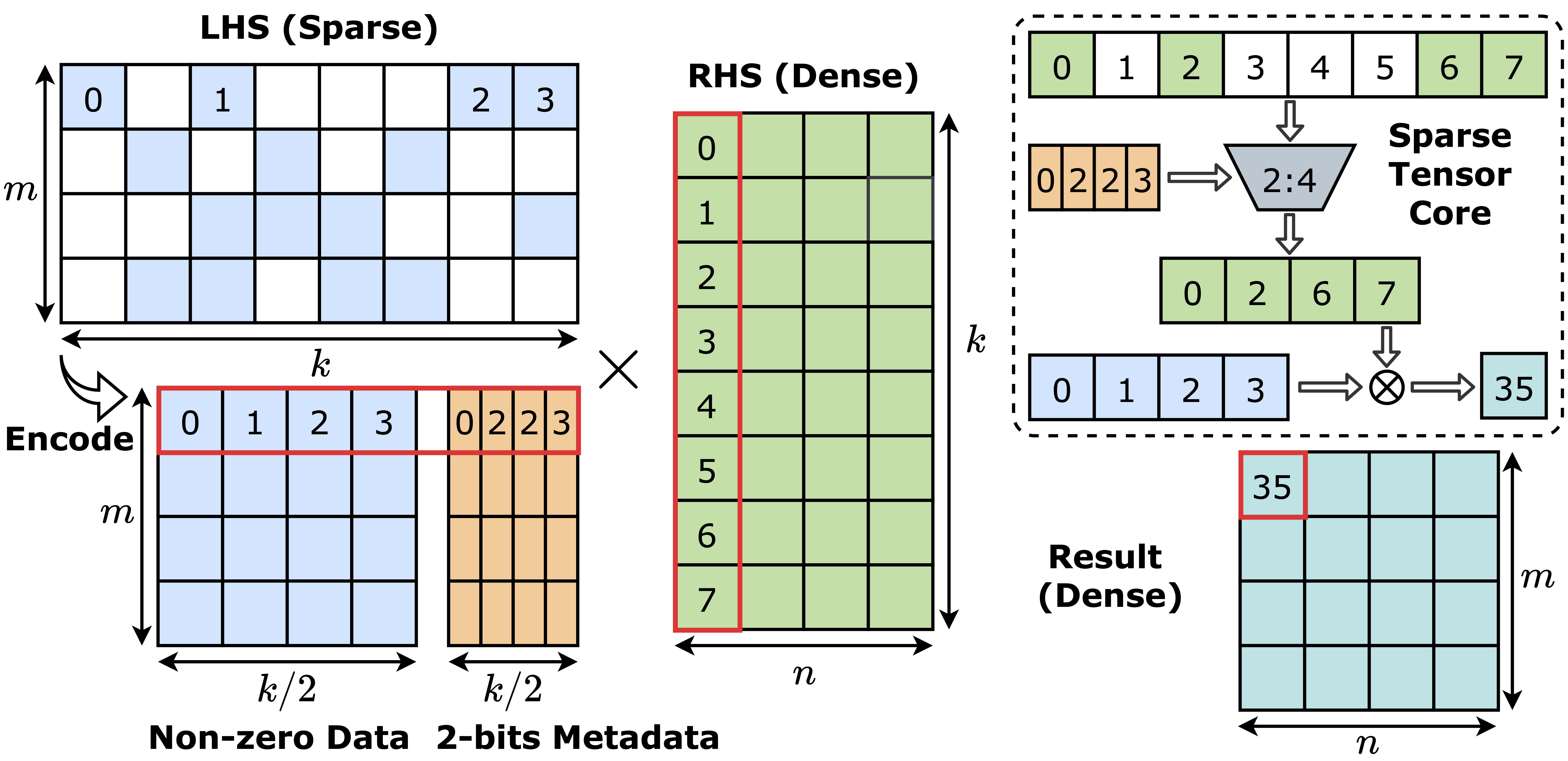}
  \caption{2:4 Sparse Encoding and Mapping for SpTC.}
  \label{fig:SpTC}
\end{figure}

\subsection{Sparse Tensor Core (SpTC)}

NVIDIA has introduced the third-generation Tensor Cores in its Ampere architecture GPUs, which can exploit the fine-grained sparsity in model parameters\cite{mishra2021accelerating}. The 2:4 sparse matrix multiplication operations are efficiently executed on SpTC, as illustrated in Figure \ref{fig:SpTC}. The original sparse matrix of size $m \times k$ is encoded into a non-zero data matrix and a 2-bits metadata matrix. The data matrix compresses all non-zero values into a dense format and the metadata matrix records the positions of non-zero elements in each contiguous set of 4 elements. NVIDIA's cuSPARSELt library provides kernels that support this format, enabling efficient compression and sparse operations\cite{cusparselt}. 
Other GPU vendors also implement similar sparse arithmetic logic unit (ALU) in their products, such as the CDNA3 series GPUs (Instinct MI300) from AMD\cite{amdcdnaISA}. 
Additionally, modern deep learning frameworks such as PyTorch\cite{PyTorchSemiSparse} and compilers like LLVM\cite{LLVM2to4} now support specific data types, operators, and intermediate representations for 2:4 sparsity, further illustrating its widespread application. Besides existing kernels, programmers can utilize the SpTC in CUDA using the \textit{mma.sp} instruction provided by Parallel Thread Execution (PTX) Instruction Set Architecture (ISA) since version 7.0, which provides the flexibility to customize kernels for various formats.

\section{Motivation}\label{sec:Motivation}

\begin{figure}[t]
  \centering
  \includegraphics[width=0.98\linewidth]{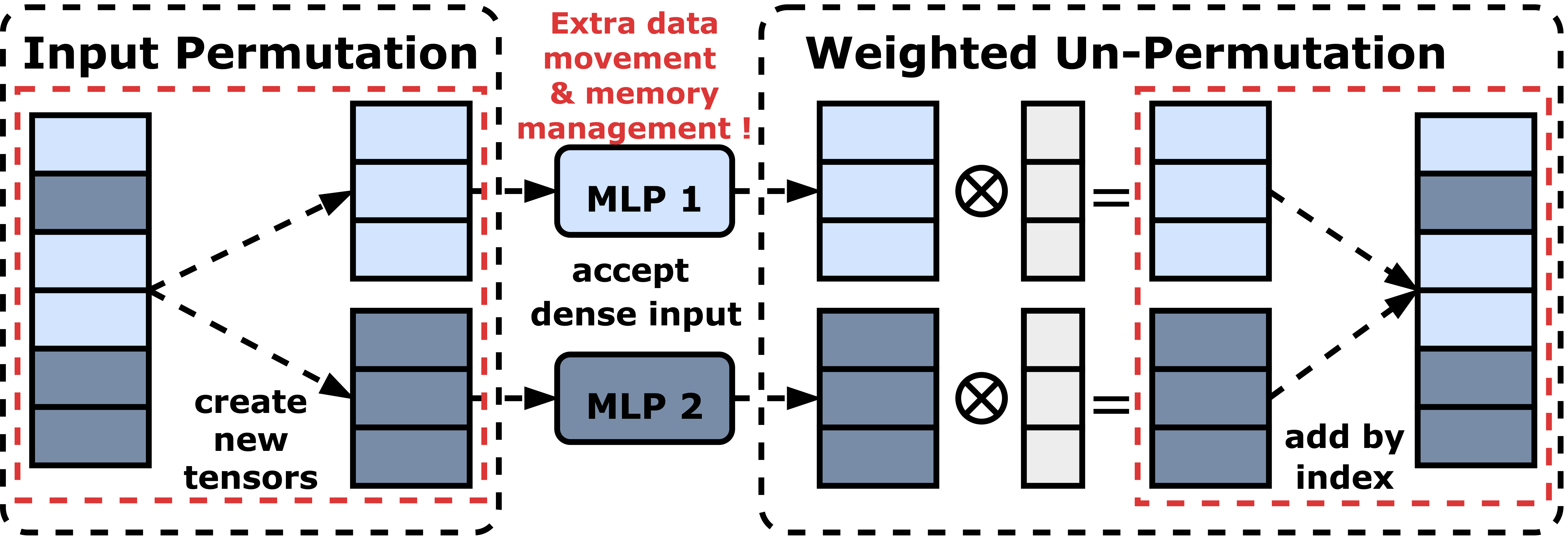}
  \caption{Redundancy in Data Flow of the MoE layer.}
  \label{fig:额外开销}
\end{figure}

\subsection{Redundancy in Data Flow}\label{sec:MoE_sparse_input}

In the MoE layer, input tokens $I$ are routed and subsequently assigned to a subset of available experts. As illustrated in Figure \ref{fig:额外开销}, the initial input tensor is permuted into several new tensors, each corresponding to a certain expert. Specifically, the input tensor for expert $E_i$ contains all the tokens that have been routed to $E_i$. The output of expert $E_i$ is then propagated to a new tensor that matches the size of $I$, ensuring that the dimension of the MoE layer output aligns with the dimension of the original input.

During the input permutation phase, the creation of extra intermediate tensors necessitates additional memory management tasks, including \textit{memory allocation} and \textit{data movement}, thereby increasing the processing overhead. In the weighted un-permutation phase, the outputs of the experts are initially transferred from registers to the global memory on GPUs. Subsequently, the element-wise product operation reloads these outputs from global memory back into the registers. This \textit{additional memory I/O} within GPUs introduces significant overheads, impacting overall performance efficiency. Therefore, these overheads necessitate a customized kernel capable of efficiently handling the sparsity in inputs.

\subsection{Redundancy in Model Parameters}\label{sec:MoE_sparse_weight}

Recent research\cite{DalviSDB20, ChenNRWS23, FrantarA23} has demonstrated redundancy in model parameters. In response, researchers have developed various sparse formats to accelerate the computing process by omitting the computation of redundant parameters.

With unstructured sparse formats like COO and CSR, individual parameters are removed based on their importance. Existing GPU kernels designed for these unstructured formats are primarily optimized for high-performance computing (HPC) applications, where the sparsity ratio often exceeds 95\%\cite{cusparse, ChenQLDX21}. In contrast, sparsity ratios in LLMs typically fall between 50\% and 90\%\cite{nmSparse}. Consequently, employing these unstructured formats in LLMs does not guarantee performance improvement, as the computational savings are often mitigated by the substantial overhead of decoding\cite{mishra2021accelerating}. With structured sparse format, parameters in different patterns are removed by group. While this format provides significant performance improvements in computation, it also limits the expressiveness of the model and reduces accuracy.

Fortunately, the N:M structured sparse format provides both the benefits of these two formats. It has been demonstrated to have a negligible impact on the accuracy of LLMs with 50\% or even larger sparse ratio\cite{CaoZYXNZLWZ19, mishra2021accelerating, ChenQQ0DX23, FrantarA23}. Meanwhile, hardware manufacturers like NVIDIA also provide hardware units (e.g. SpTC) to further accelerate the computation process. Therefore, to reduce the cost of computing redundant model parameters, kernels should be capable of handling this N:M structured sparse format.

\subsection{Problem with Existing Solutions} \label{sec:motivation_existing_problem}

\begin{figure}[t]
    \centering
    \includegraphics[width=0.93\linewidth]{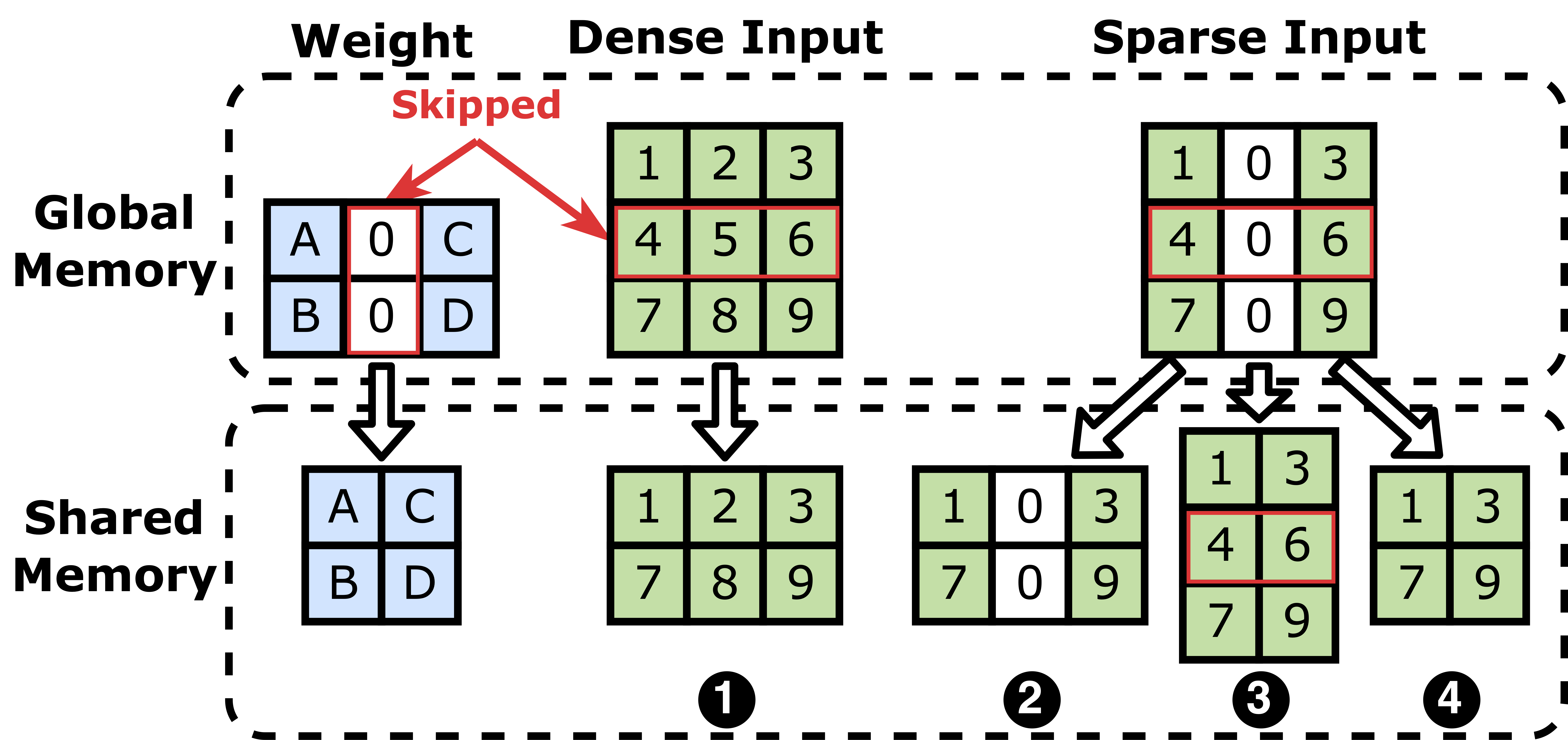}
    \caption{Inefficient Memory Access when Input is Sparse. \ding{202} illustrate the dense scenario. \ding{203} and \ding{204} lead to I/O amplification, while \ding{205} causes uncoalesced memory access.}
    \label{fig:venom缺陷}
\end{figure}

To address the aforementioned redundancy in MoE execution, several solutions\cite{megablocks,vllm,VENOM} have been proposed. However, these works either fail to explore the the potential of structured sparsity in model parameters, which can be accelerated by sparse ALU, or waste the memory bandwidth for dual-side sparse matrix multiplication.

Among them, Megablocks\cite{megablocks} provides a block sparse representation and a customized kernel, while vLLM\cite{vllm} proposes a kernel that combines the computation processes of all experts into a single kernel to address data flow redundancies in MoE layers. However, these solutions overlook the opportunity to leverage structured sparsity in model weights. Moreover, their highly customized designs make it challenging to incorporate the structured sparsity efficiently.

Meanwhile, other research has developed kernels optimized for structured sparsity, delivering notable performance improvements over SOTA kernels for dense matrices or unstructured sparse matrices. However, kernels like BBS\cite{CaoZYXNZLWZ19} and nmSPARSE\cite{nmSparse} fail to utilize SpTC for further acceleration. Solutions such as cuSPARSELt\cite{cusparselt} and DFSS\cite{ChenQQ0DX23} leverage SpTC but impose a sparse ratio limit of 50\%. VENOM\cite{VENOM} allows for a flexible sparse ratio while utilizing SpTC with a V:N:M format, specifically optimized for sparse-dense matrix multiplication scenarios. As depicted in Figure \ref{fig:venom缺陷}, when encountering a sparse column in model weights, it skips the multiplication with the corresponding row in inputs. This approach maintains an efficient memory access pattern with coalesced memory access, as illustrated in \ding{202}. 

However, challenges arise when both weight and input matrices are sparse. In such situations, as shown in Figure \ref{fig:venom缺陷}, the skipped row and the sparse column in inputs break the data into smaller tiles, adversely reducing performance. For instance, the data may be loaded into shared memory in formats \ding{203}, \ding{204}, and \ding{205}. Formats \ding{203} and \ding{204} involve loading either sparse column data or skipped row data unnecessarily, leading to severe I/O amplification at high sparse ratios. Moreover, the data in format \ding{205} are not contiguous in memory, leading to uncoalesced memory access and reducing GPU memory I/O bandwidth.

\section{Design of Samoyeds} \label{sec:design}

To address the limitations outlined in \S\ref{sec:Motivation}, Samoyeds introduces a new sparse data format and a dedicated kernel execution scheme. Furthermore, we adopt novel systematic optimizations, including tiling, data stationary, packing, and optimized layout, specifically tailored for efficient structured sparse MoE computation.

\begin{figure}[t]
  \centering
  \includegraphics[width=0.95\linewidth]{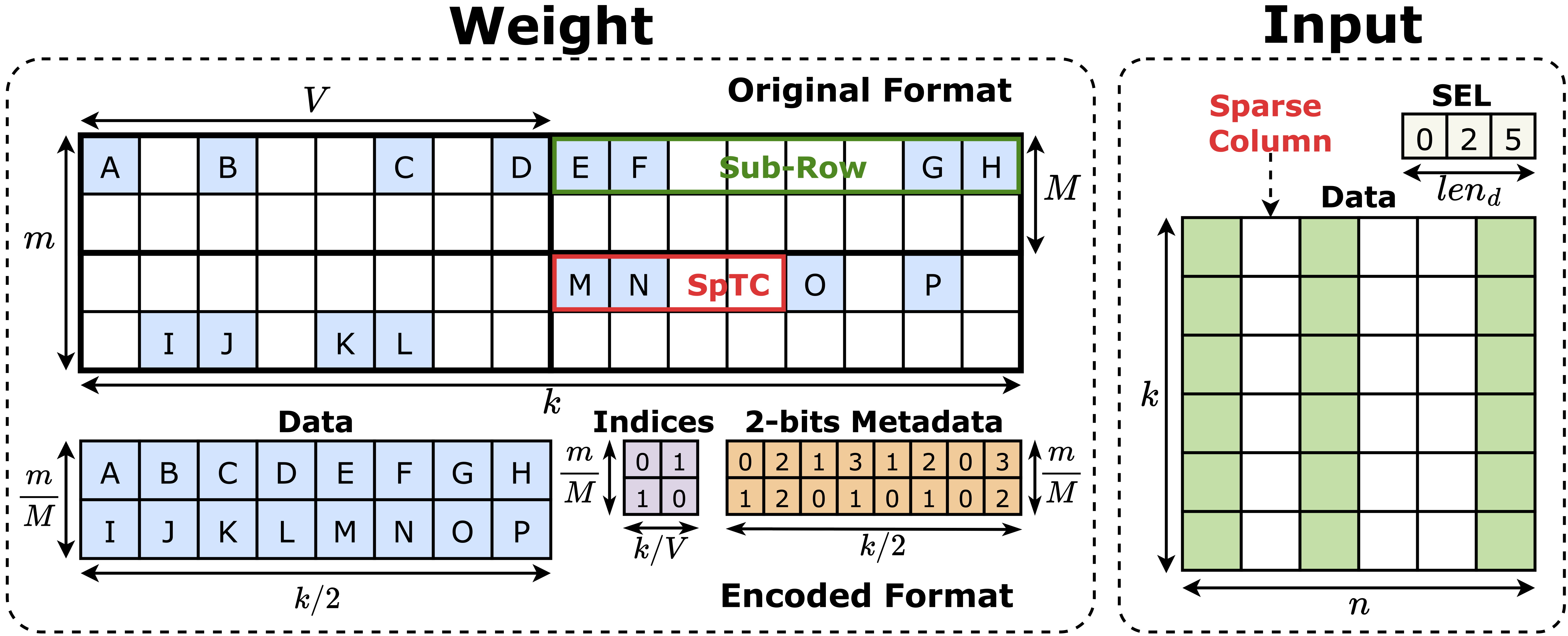}
  \caption{Samoyeds Dual-side Sparse Data Format.}
  \label{fig:稀疏表示}
\end{figure}

\subsection{Dual-sided Sparse Data Format and Kernel Scheme} \label{sec:sparse_encoding}

As discussed in \S\ref{sec:motivation_existing_problem}, existing structured sparse data format can result in uncoalesced memory access or memory I/O amplification, thereby degrading overall kernel performance. Samoyeds firstly introduces a novel structured sparse format specifically tailored for sparse-sparse matrix multiplication in MoE computation, employing distinct sparsity patterns for weight and input matrices, as illustrated in Figure \ref{fig:稀疏表示}.

For weight matrices, the sparse pattern integrates 2:4 element-wise and vector-wise structured sparsity. Samoyeds employs the 2:4 pattern to align with the ISA requirements of the SpTC, leveraging the superior speedup of sparse ALU. Given that the 2:4 pattern enforces a fixed sparsity ratio of 50\%, an additional sparse pattern is required to provide greater flexibility. 
The granularity of the added sparse pattern should be more coarse-grained to preserve the prior 2:4 element-wise sparse pattern.
Prior studies\cite{nmSparse} have proven block-wise sparsity is too coarse-grained to preserve model accuracy. Therefore, Samoyeds adopts vector-wise sparsity alongside the existing 2:4 sparsity, striking a balance between representation flexibility and accuracy preservation.

The weight sparse pattern is shown on the left of Figure \ref{fig:稀疏表示}. The original weight data is segmented into structured sparse blocks of size $M \times V$. Each vector within a block, termed as a \textit{Sub-Row}, contains multiple \textit{SpTC units}. Within each block, only N Sub-Rows are retained, while the others are pruned, where N depends on the target sparse ratio. The SpTC units within the selected Sub-Row are further pruned to conform to the sparse pattern supported by the hardware.

Based on the defined sparse pattern, the original weight is encoded into three components: data, indices, and metadata. The data matrix, with the shape of $\frac{m}{M} \times \frac{k}{2}$, serves as a compressed representation of the original sparse matrix of size $m \times k$. The compression ensures that elements are ordered sequentially, enabling GPU-friendly decoding during computation. The indices matrix, with size $\frac{m}{M} \times \frac{k}{V}$, captures the relative positions of the retained Sub-Rows within their respective blocks. Meanwhile, the metadata matrix, with size $\frac{m}{M} \times \frac{k}{2}$, details the sparse pattern for each SpTC unit. Notably, each item within the metadata matrix is encoded into 2-bits as required by the SpTC hardware specifications.

The input data sparse pattern, illustrated on the right of Figure \ref{fig:稀疏表示}, introduces a selection array (\textit{SEL}) and encodes the columns in vector-wise sparsity way. Notably, this design naturally aligns with the sparsity pattern presented in token routing, ensuring mathematical equivalence with the original computation process.

\begin{algorithm}[t]\scriptsize
    \caption{Samoyeds Kernel Scheme}\label{alg:kernel}
    \tikzmk{A}
        \tcp{Inputs in Samoyeds sparse format. \quad \quad \quad  // Optimized Layout}
        \KwIn{\textit{A}, \textit{Indices}, \textit{Metadata}, \textit{B}, \textit{SEL}}
    \tikzmk{B}\boxit{mypink}{0.005}{0.1}{0.92}{0.68}
    \KwOut{\textit{C}}

    Init shared memory for \textit{A}, \textit{Indices}, \textit{B}, \textit{SEL}\;
    Init registers for \textit{A}, \textit{Indices}, \textit{Metadata}, \textit{B}, \textit{SEL}, \textit{C}\;
    Load All \textit{SEL} from GMEM to SMEM\;
    $fetch \gets 0$\;
    
    \For{$compute = 0$ to $\frac{k}{k_b}$}{
        \tikzmk{AA}
            Load \textit{Metadata} from GMEM to Register \tcp*{Packing}
        \tikzmk{BB}\subboxit{myblue}{0.005}{0.2}{0.85}{0.8}     
        \While{$fetch < compute + num_{pipe}$ \textbf{and} $fetch < \frac{k}{k_b}$}{
            \tikzmk{A}
                \tcp{Fetch Stage}
                \tikzmk{AA}
                    Load \textit{I}, \textit{A}, \textit{B} from GMEM to SMEM \tcp*{Tiling}
                \tikzmk{BB}\subboxit{mygreen}{0.01}{0.2}{0.78}{0.8}
                Commit group for pipeline\;
                Increase $fetch$;
            \tikzmk{B}\boxit{mygrey}{-0.00}{0.1}{0.59}{-0.3}
        }
        
        Wait group for pipeline\;
        \tikzmk{A}
            \tcp{Compute Stage}
            \tikzmk{AA}
                Load \textit{I}, \textit{A}, \textit{B} from SMEM to Register \tcp*{Tiling}
            \tikzmk{BB}\subboxit{mygreen}{0.01}{0.2}{0.84}{0.8}
            \tikzmk{AA}
                \If{$compute \equiv 0 \pmod{\frac{V}{k_b}}$}{Shuffle \textit{C} Register \tcp*{Data Stationary}}
            \tikzmk{BB}\subboxit{myyellow}{0.01}{0.1}{0.84}{0.8}
            Sparse MMA computations;
        \tikzmk{B}\boxit{mygrey}{-0.00}{0.1}{0.513}{-0.3}
    }
    \tikzmk{A}
        Store \textit{C} from Register to GMEM \tcp*[f]{Optimized Layout}
    \tikzmk{B}\boxit{mypink}{0.005}{-0.35}{0.015}{0.2}
\end{algorithm}

To unleash the potential of the proposed sparse data format, Samoyeds customizes a GPU kernel to accelerate computation. The pseudo-code of the kernel is described in Algorithm \ref{alg:kernel}. After receiving the encoded sparse input, the kernel begins with an initialization phase, followed by the fetch stage (line 10-13), where data is loaded from global memory to shared memory using the non-blocking \textit{cp.async} instruction. This asynchronous copy operation is then committed in group. Next, in the compute stage (line 16-21), the kernel invokes the wait method for a specific group and synchronously loads data from shared memory into registers according to the predefined tiling size. Once all data are in place, the \textit{mma.sp} instruction triggers SpTC to perform sparse computation. These two stages (fetch and compute) are efficiently overlapped using a pipeline mechanism. At the end of the kernel execution, the output matrix $C$ is transferred from the registers back to global memory. 

Notably, while our kernel scheme aligns with the typical execution flow of matrix multiplication kernel, adapting it to the unique Samoyeds data format and orchestrating various kernel optimization techniques—such as tiling, data stationary, packing, and optimized layout—is a non-trivial task. The following sections provide a detailed description of the optimizations implemented in Samoyeds to enhance performance.

\begin{figure}[t]
  \centering
  \includegraphics[width=0.95\linewidth]{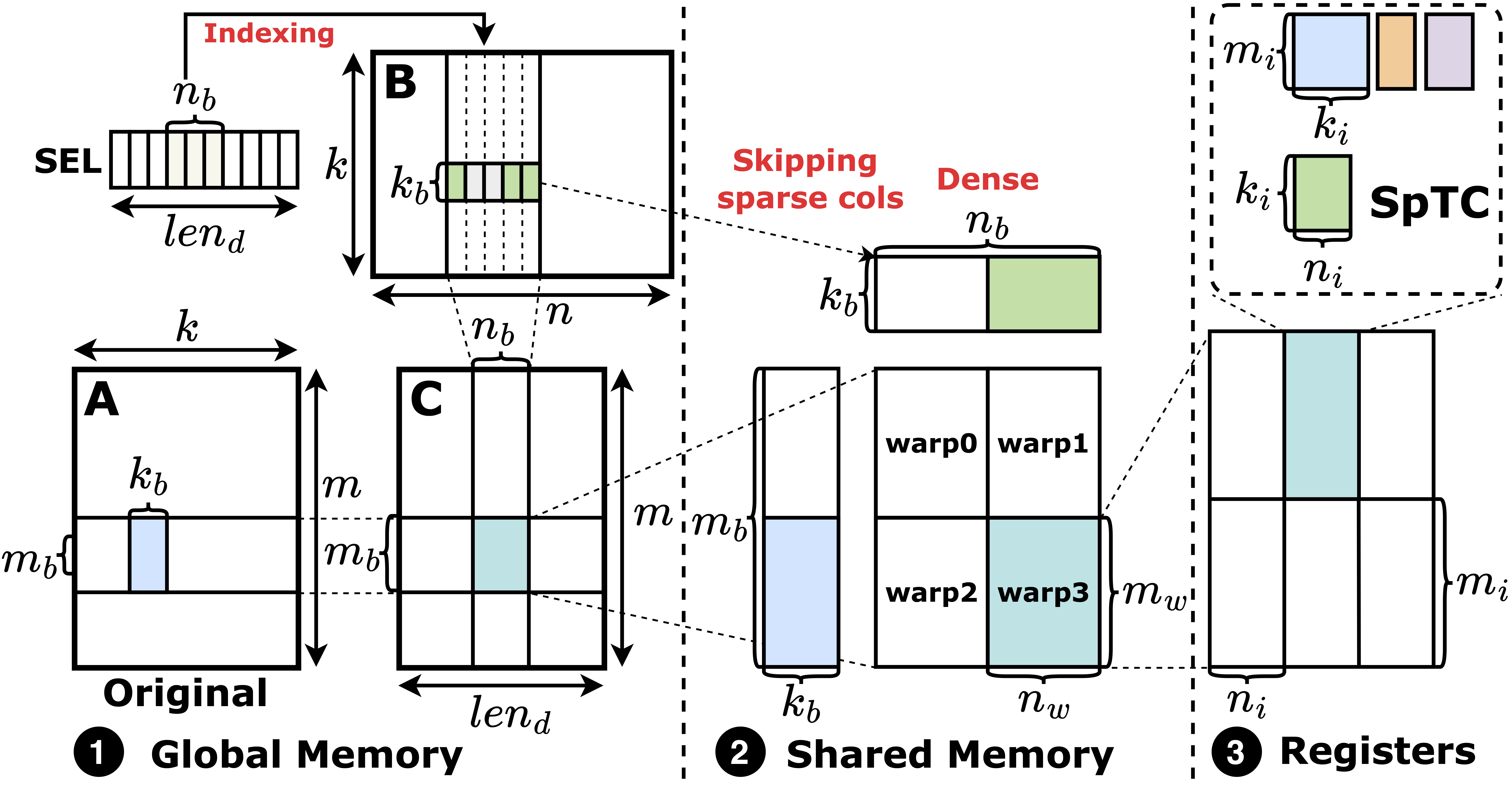}
  \caption{Tiling Strategy for GPU Memory Hierarchy.}
  \label{fig:Tiling}
\end{figure}

\subsection{Enhanced Data Locality with Tiling} \label{sec:tiling_design}

Tiling is a widely used technique in GPU kernel design that partitions data into equal-sized subsets to exploit data locality. 
By leveraging the multi-level memory hierarchy of GPUs, this approach significantly enhances memory access efficiency, reducing data-loading latency and increasing computational throughput.
However, the Samoyeds data format encodes the original matrix into multiple matrices, fundamentally altering the data access patterns. Therefore, orchestrating an effective tiling strategy for these matrices becomes a non-trivial task.

As illustrated in Figure \ref{fig:Tiling}, we introduce a three-step tiling strategy to optimize Samoyeds kernel. Considering a $m\times k\times n$ matrix multiply problem, denoted as $C=A\times B$. Notably, only $len_d$ columns from matrix B are selected for computation, which are recorded in the selection array. For clarity, matrix A is presented in its original, non-encoded format. In step \ding{202}, each thread block is responsible for computing a tile of size $m_b \times n_b$ in matrix C. During each iteration, data in the global memory is further partitioned along the $k$ dimension. Each thread block then loads segments of matrices A ($m_b \times k_b$), B ($k_b \times n_b$) into shared memory. Note that although the tiling size for matrix B specifies $n_b$ columns, the original layout in global memory includes more columns than $n_b$ due to the omission of sparse columns. Only the columns identified by the mapping in the selection array are actually loaded, ensuring an efficient use of memory resources. 
In step \ding{203}, the output of matrix C is segmented and assigned to several warps, with each warp handling a tile size of $m_w \times n_w$. Given that threads in modern GPUs are organized into scheduling units called warps, each warp loads the corresponding sections of matrices A and B from shared memory into thread registers. In step \ding{204}, these registers are further partitioned to align with the requirements of the SpTC hardware. The SpTC is then invoked to compute the output size of $m_i \times n_i$ for acceleration.
Moreover, since the metadata matrix, encoded in 2 bits, is relatively small, applying 3-step tiling leads to inefficient memory access on GPUs. Therefore, it skips the innermost tiling step and is loaded directly into registers.
Similarly, the indices matrix is loaded using the 3-step tiling scheme but with a larger tiling size than the corresponding matrix 
$A$ to enhance memory efficiency.

The selection of the tiling size requires careful consideration of multiple factors from various perspectives. 
First, hardware specifications impose constraints on the tiling size. Specifically, the values of $(m_i, k_i, n_i)$ must comply with the requirements of the \textit{mma.sp} instruction. Additionally, in the three-step tiling scheme, the amount of data loaded into shared memory and registers is bounded by available hardware resources.
Second, while a larger tiling size improves data locality during kernel execution, a smaller tiling size increases parallelism by dividing the matrix into more execution units, thereby enhancing GPU hardware utilization. Consequently, a trade-off must be carefully balanced based on the specific problem size.
Third, when integrated into MoE models, an excessively large subrow size $V$ may degrade model accuracy. Additionally, the tiling size $K_b$ for the reduction dimension $K$ is constrained by $V$. Therefore, we employ a larger tile size in non-reduction dimensions ($M$ and $N$) to improve data reuse while keeping $K_b$ relatively small to avoid accuracy loss. Additionally, for models with more experts, the tiling size for $N$ dimension should be reduced to avoid potential padding overhead or degraded hardware efficiency.

\begin{figure}[t]
    \centering
    \subfigure[Problem Illustration.]{
        \includegraphics[width=0.30\linewidth]{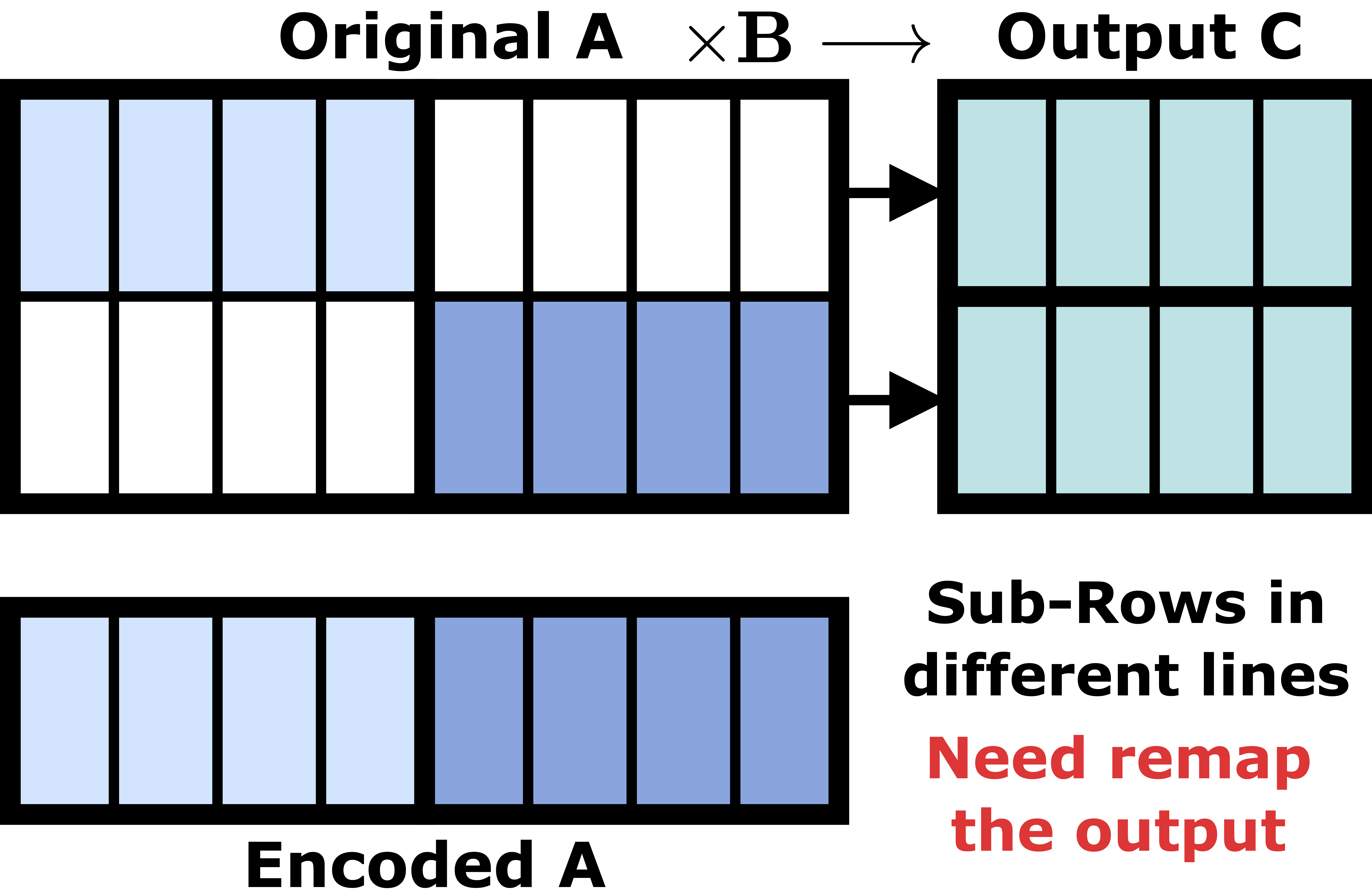}
        \label{fig:Stationary_Problem}
    }
    \centering
    \subfigure[Introducing Intermediate Registers.]{
        \includegraphics[width=0.6\linewidth]{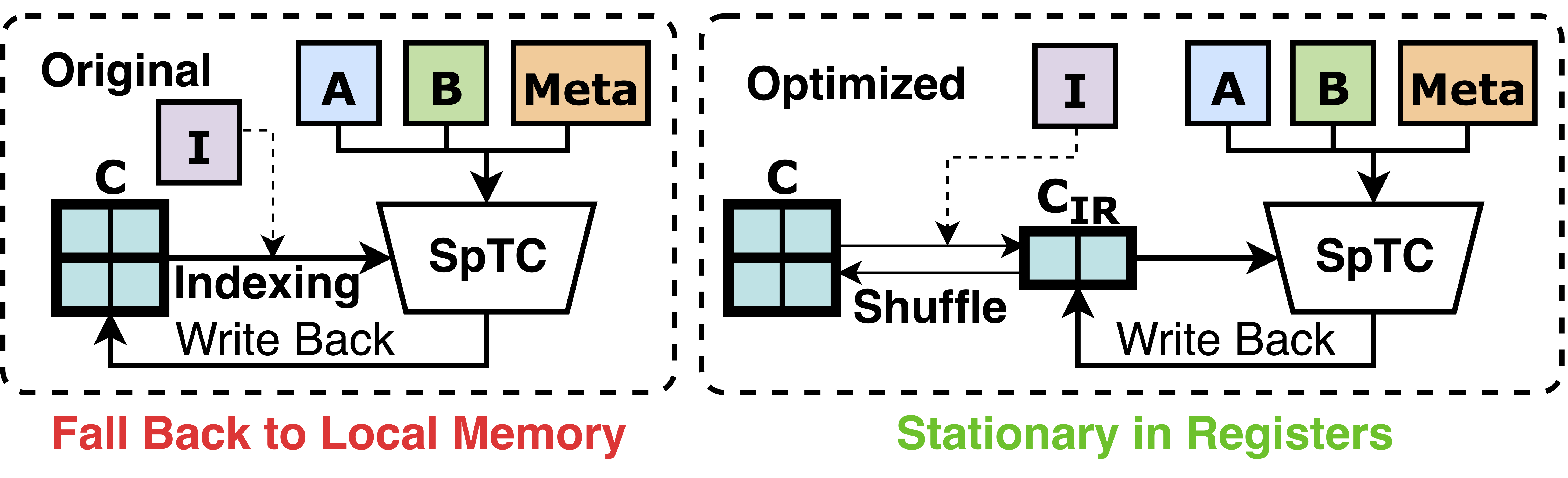}
        \label{fig:Stationary_Solution}
    }
    \caption{Data Stationary Optimization for Output Matrix.}
\end{figure}

\subsection{Maximized Data Reuse with Data Stationary } \label{sec:data_stationary}
Data stationary refers to the strategy of pinning data elements in faster memory hierarchy throughout the tiling loops. However, given that Samoyeds encodes data in a unique format, directly applying existing designs can result in suboptimal choices for stationary locations, potentially resulting in degraded kernel performance.

In matrix multiplication, the input matrices $A$ and $B$ are read-only. In contrast, the output matrix $C$ requires both reading and writing in each tiling iteration. Thus, a well-established design approach\cite{BLIS1} is keeping the input data in shared memory and maintaining the output data in registers.

However, when iterating on reduction along the $k$ dimension, the Sub-Rows may span across different lines between sparse blocks, as illustrated in Figure \ref{fig:Stationary_Problem}. This presents a new challenge: when the tiling window shifts from one Sub-Row to another, the output of the SpTC must be remapped to different rows. Thus, when invoking the \textit{mma.sp} instruction, it is crucial to carefully select the output registers based on the mappings recorded in the indices matrix to ensure the correctness of the computation. However, simply passing the indexed output to the instruction can result in the stationary location of output $C$ falling back to local memory on GPU, as shown on the left of Figure \ref{fig:Stationary_Problem}, which will significantly reduce the kernel performance. 

To address this challenge, we introduce additional intermediate registers $C_{IR}$, as shown on the right of Figure \ref{fig:Stationary_Solution}. All registers storing $C$ are initialized to zero at the beginning of kernel execution. Stored $C$ data is shuffled with $C_{IR}$ according to the indices matrix every $\frac{V}{k_b}$ iteration, which corresponds to the point when the tiling window shifts from one Sub-Row to another. This optimization minimizes frequent memory transfers between global memory and registers, thereby enhancing overall kernel performance with mathematical correctness.

Besides, Samoyeds adopts operator fusion to further enhance data reuse by improving cache utilization, eliminating intermediate results, increasing locality, and reducing data movement. This technique allows the succeeding operator to utilize the results of the preceding operator without roundtrips to global memory. Specifically, the activation function and its precedent operator are fused. Furthermore, the weighted accumulation operation, which contains a broadcast of scalar data and dot multiplication, is fused with matrix multiplication.

\begin{figure}[t]
  \centering
  \includegraphics[width=\linewidth]{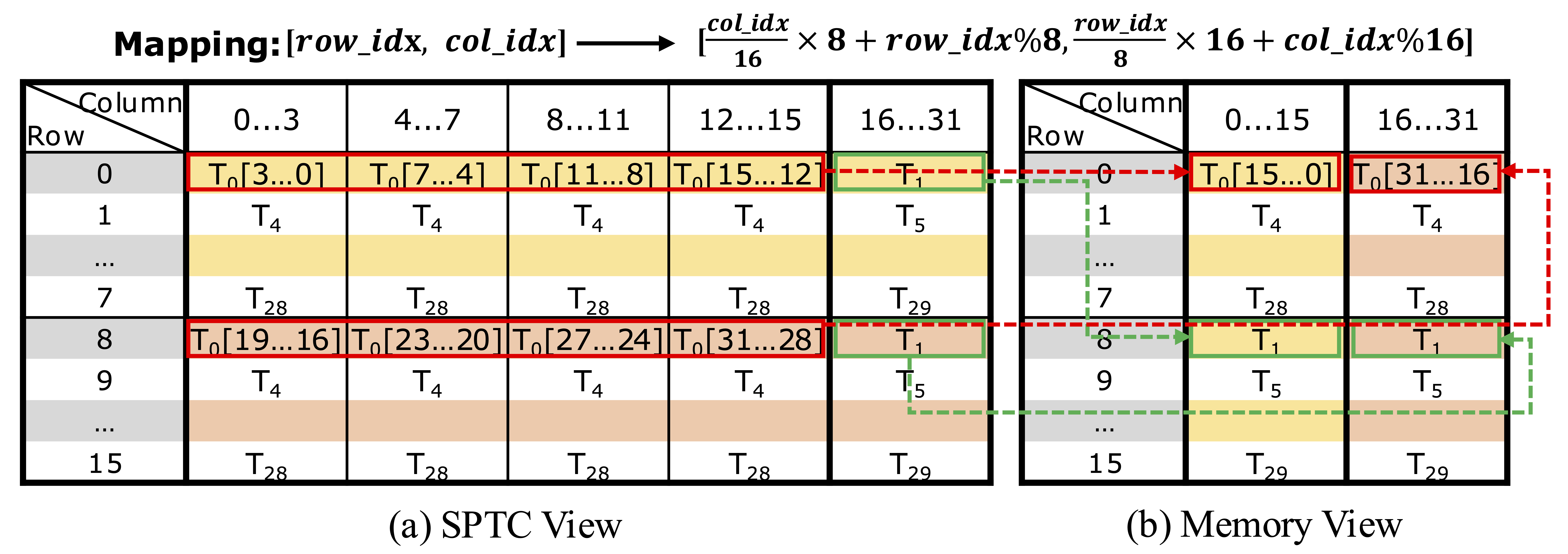}
  \caption{Packing Strategy with Reorganized Metadata.}
  \label{fig:metadata_packing}
\end{figure}

\subsection{Coalesced Memory Access with Data Packing}

Data packing is a commonly used technique to enhance kernel performance by enabling coalesced memory access and maximizing memory bandwidth. However, the effectiveness of this technique is closely tied to the data representation format. Our newly proposed Samoyeds sparse format differs significantly from existing sparse formats, posing challenges for directly integrating existing optimizations into the Samoyeds kernel. Additionally, the data must be organized to align with the SpTC specifications, requiring customized packing designs. Samoyeds introduces a data packing strategy that optimally arranges the storage of data matrices $A$ and $B$, as well as metadata.

According to the SpTC specification, the data of each thread is not contiguous in input matrices. For matrix $A$, data is packed in global memory according to the matrix format, and the transfer from global memory to shared memory conforms to the 128-bit memory transactions of modern GPUs. Compliance with the SpTC specification is achieved during transferring from shared memory to registers by invoking \textit{ldmatrix} instructions. Additionally, data in shared memory is arranged with permutation to prevent bank conflicts. Considering matrix $B$, which retains row contiguity and exhibits column sparsity, we pack it with transposition in global memory. This transposition allows for contiguous memory access within rows and enables skipping over rows with zero values, optimizing memory bandwidth utilization through coalesced memory access. The approach for packing matrix $B$ in shared memory and transferring data to registers mirrors that of matrix $A$.

However, for the metadata matrix, each element is encoded into the 2-bits format, incompatible with the specification of \textit{ldmatrix} instruction. For efficient memory access, we propose a special packing format for the metadata matrix. Taking the \textit{mma.sp.m16n8k32} instruction for example, the metadata in bfloat type occupies a 32-bits register containing 16 2-bits vectors from the view of SpTC, as shown in Figure \ref{fig:metadata_packing}(a). We need to concatenate four consecutive metadata into a 16-bit metadata block and iterate twice to fill the 32-bits register. To avoid missing in the L2 cache, our proposed metadata packing for continuous access on device memory is shown in Figure \ref{fig:metadata_packing}(b). To be specific, the metadata is a 2-bits matrix with a size of 16$\times$16. The element with location of [$row\_idx, col\_idx$] is mapped to the location of [$row\_idx \% 8 \times 2 + \frac{col\_idx}{8}, col\_idx \% 8 + \frac{row\_idx}{8} \times 8$]. Having got the new data mapping, the metadata loading for each thread is aligned to 32-bits, which is consistent with the 32-bits memory transactions of GPUs. 

\subsection{Reduced Memory I/O with Optimized Layout}\label{sec:data_layout_opti}

The data layout is crucial for optimizing GPU kernels, as an efficient layout reduces memory I/O and enhances computational performance. However, the constraints imposed by hardware instructions, combined with the sparsity inherent to the MoE computation, introduce additional challenges for data layout design. These limitations make existing solutions inefficient in such scenarios, necessitating specialized optimizations in the data layout.

Traditionally, the linear layer performs the operation $xW$, where $x$ represents the input and $W$ denotes the model weights. To align with SpTC hardware requirements, this computation is restructured to $(W^T x^T)^T$, where $T$ denotes the matrix transpose operation. This transformation can significantly increase I/O volume across the memory hierarchy of GPUs. To mitigate this overhead, Samoyeds employs the three-step layout optimization for operands of kernels. Firstly, the transposition of the $W$ matrix is performed during the offline model pruning phase, eliminating the memory I/O of transposition. Secondly, the input $x$ for the MoE layer typically shows sparsity by row, corresponding to the token routing results, which allows for efficient contiguous storage in row-major format. Transposing this input prior to matrix multiplication would result in inefficient, scattered memory accesses, amplifying memory I/O. To address this, the transposition of the input is efficiently executed within the kernel as data transfers from global to shared memory, leveraging the hardware fast path feature of GPUs. Besides, the output transposition is integrated within the kernel, further minimizing memory I/O and boosting computational efficiency.

\begin{figure}[t]
    \centering
    \subfigure[Output Sparsity in MoE Layer.]{
        {\includegraphics[height=0.085\textheight]{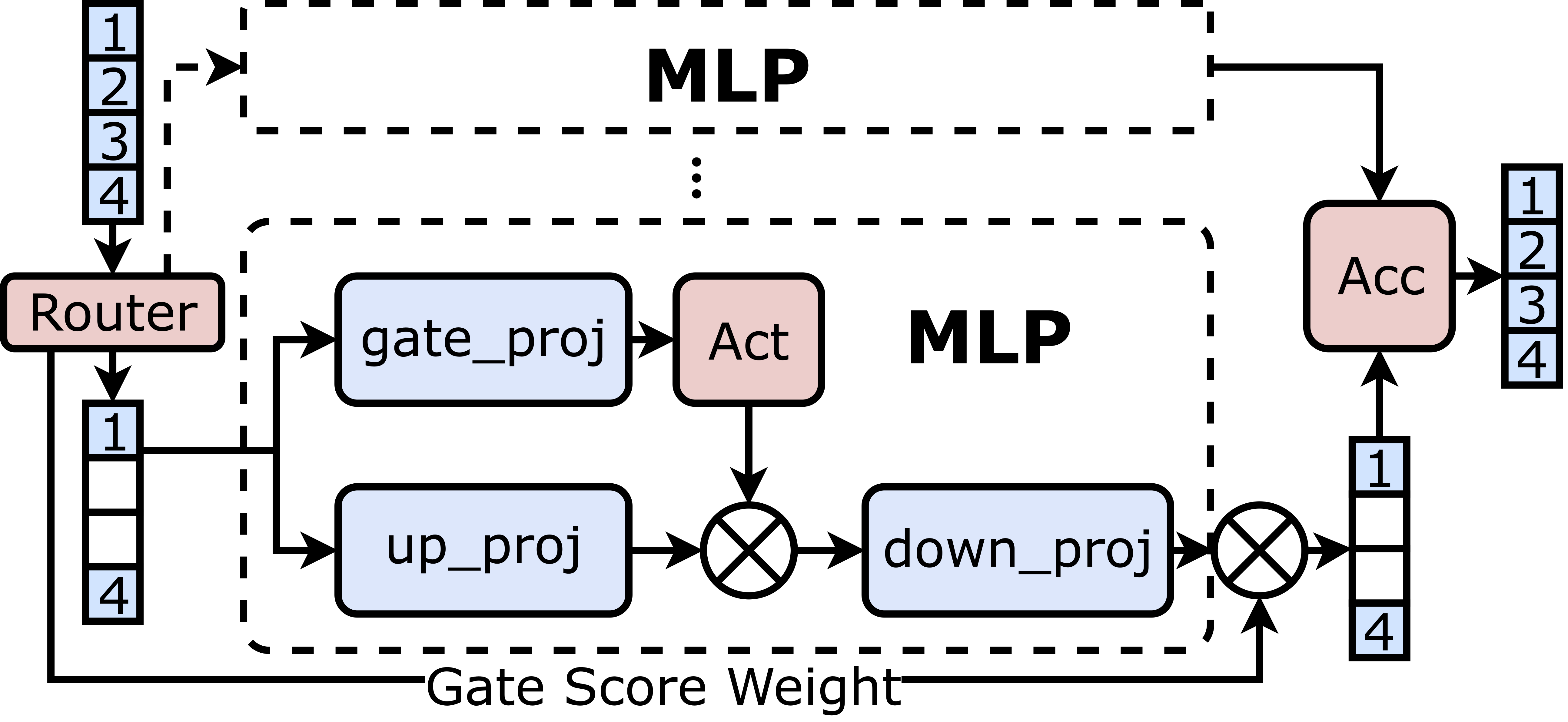}}
        \label{fig:Linear_repalcing}
    }
    \centering
    \subfigure[Kernel Performance Improvement Optimized Output Layout.]{
        \includegraphics[height=0.109\textheight]{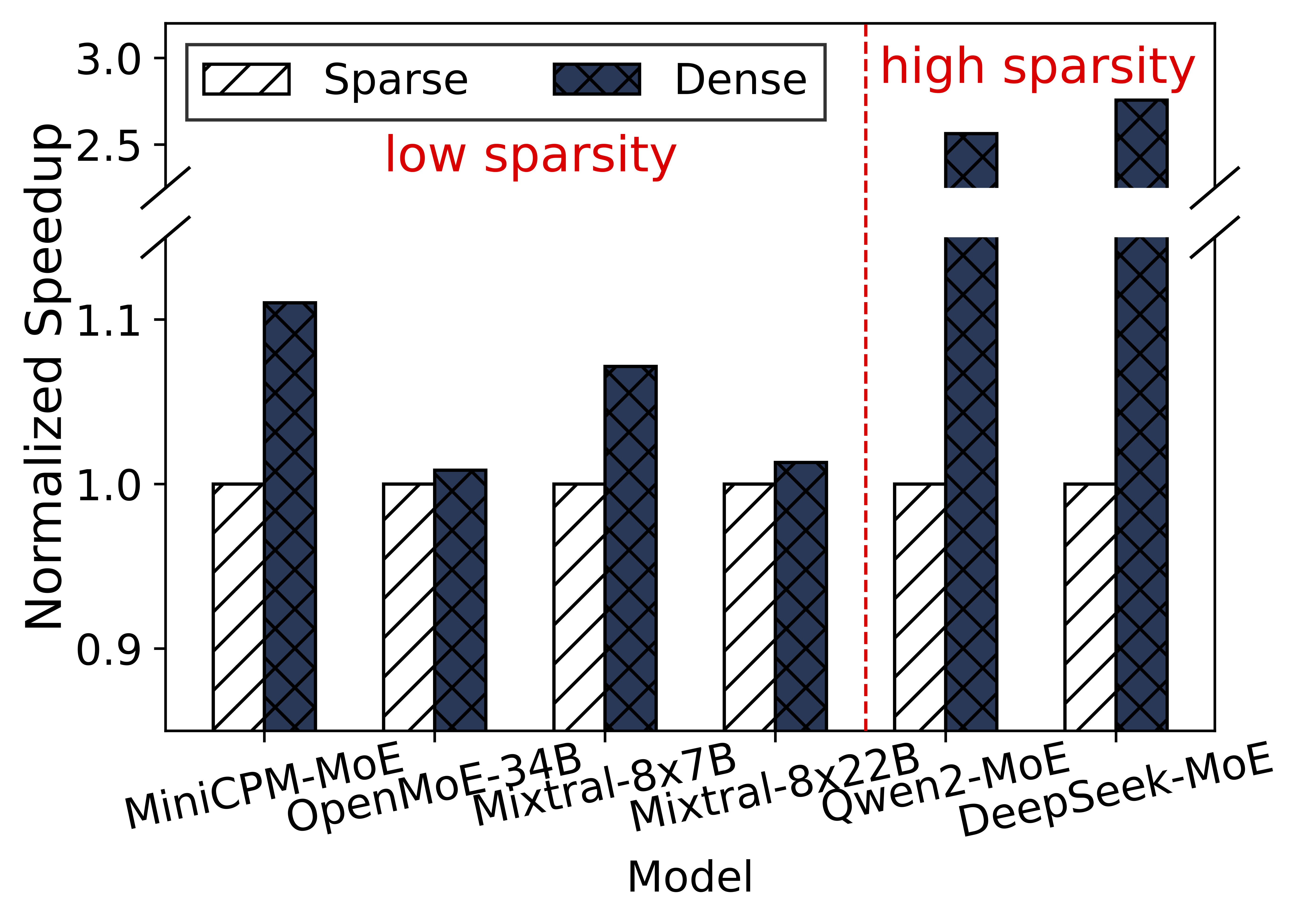} 
        \label{fig:kernel_output_format}
    }
    \caption{Layout Optimization for Kernel Output.}
\end{figure}

As shown in Figure \ref{fig:Linear_repalcing}, a typical expert layer consists of three linear layers: \textit{gate\_proj}, \textit{up\_proj}, and \textit{down\_proj}. The token routing mechanism selects a subset of the input for each expert and aggregates the outputs from all experts with weighted accumulation (\textit{Acc}). While the final output of the MoE module is dense after accumulation, the intermediate results within each expert exhibit a row-wise sparse pattern, with the sparsity ratio determined by the number of experts. This sparsity can lead to unnecessary memory transfers of zero values, causing performance degradation. To address this issue, Samoyeds adopts a compressed output layout that aligns with the input sparse pattern described in \S\ref{sec:sparse_data_formats}. This optimized layout eliminates redundant data transfers while preserving the mathematical equivalence of the computation. 
The performance gain achieved with this layout across varying input sparsity ratios is demonstrated in Figure \ref{fig:kernel_output_format}. 
For typical MoE model configurations, this optimization speeds up the kernel by 1.05$\times$ on average for models with low input sparsity and 2.66× for models with higher sparsity.

\section{Implementation}

\subsection{Kernel Implementation}

The Samoyeds kernels are implemented in CUDA, utilizing the SpTC hardware with inline assembly \textit{mma.sp} instructions from NVIDIA PTX ISA. These kernels are compiled into a dynamic library via the NVIDIA CUDA Compiler (NVCC), making them accessible to other programs. Additionally, the compiled executable is exposed as a Python module through module registration with pybind11.

\subsection{Compatibility with Different Hardware}

The mandatory requirement of the Samoyeds kernel is sparse ALU, which is specifically designed for matrix multiplications involving a sparse matrix with 2:4 structured sparsity and a dense matrix. Beyond the computational units, memory efficiency can also influence kernel performance. On the one hand, the pipeline mechanism, foundational to this implementation, leverages asynchronous data movement (e.g. cp.async) and concurrent kernel execution to enhance throughput. On the other hand, the efficiency of kernel execution is further augmented by collective load and store (e.g. ldmatrix), which orchestrate multiple threads—such as those encapsulated by wrap in CUDA and wave in ROCm.

\begin{table}
  \caption{Hardware Support for Samoyeds on Prevalent GPUs. \xmark{$^\ast$} denotes missing direct ISA support but implementable by combining other basic instructions.}
  \label{tab:hardware.list}
  \resizebox{\linewidth}{!}{
      \begin{tabular}{ccccc}
        \toprule
        \multirow{3}{*}{GPU} & \multirow{3}{*}{Architecture} & \multicolumn{1}{c}{Mandatory} & \multicolumn{2}{c}{Optional} \\ 
        \cmidrule(lr){3-3} \cmidrule(lr){4-5}
        &  & \makecell[c]{Sparse \\ALU} & \makecell[c]{Asynchronous \\Memory Copy} & \makecell[c]{Collective \\Load/Store} \\
        \midrule
        % 3090  & NVIDIA & Ampere & \cmark  & \cmark & \cmark \\
        NVIDIA A100 & Ampere & \cmark & \cmark & \cmark \\
        NVIDIA 4090 & Ada Lovelace & \cmark & \cmark & \cmark \\
        NVIDIA H100 & Hopper & \cmark & \cmark & \cmark \\
        AMD W7900 & RDNA3  & \xmark & \xmark{$^\ast$} & \xmark{$^\ast$} \\
        AMD MI300 & CDNA3  & \cmark & \xmark{$^\ast$} & \xmark{$^\ast$} \\
      \bottomrule
      \end{tabular}
  }
\end{table}

The Samoyeds kernel can be effectively adapted to other accelerators equipped with the aforementioned features.
As demonstrated in Table \ref{tab:hardware.list}, the Samoyeds kernel is compatible with most prevalent GPUs, including NVIDIA GPUs that incorporate Ampere architecture or more advanced versions\cite{nvidiaptx}, as well as AMD GPUs that are equipped with CDNA3 architecture\cite{amdcdnaISA}, which all satisfy the mandatory requirement for sparse ALU.
However, the lack of native support for asynchronous memory copy and collective matrix load/store operations on AMD GPUs may lead to degradation in memory efficiency and increased development effort.
Despite the applicability, the optimized kernel configuration (e.g. pipeline stages and tiling size) can be distinct on different hardware, based on their difference in resources, including the number of clusters/processors, cache line size, and shared memory capacity. The relationship between kernel configuration and hardware specification will be further explored in \S\ref{sec:HardwarePortability}.

\section{Evaluation}

The performance of Samoyeds system is evaluated on 3 distinct levels, including kernel-level performance improvements (\S\ref{sec:kernel}), the enhancements achieved within the MoE layer (\S\ref{sec:MoELayer}), the speedup and batch size benefits in end-to-end scenarios (\S\ref{sec:E2E}). A breakdown analysis is provided in \S\ref{sec:ablation}. Additionally, the accuracy of models pruned with Samoyeds sparse format are assessed in \S\ref{sec:accuracy}. Finally, the performance portability is examined in \S\ref{sec:HardwarePortability}.

\textbf{Evaluation Setup.} The evaluation is conducted on the platforms equipped with Intel i7-12700 CPU, 16G$\times$2 DDR5 memory, running Ubuntu 22.04LTS and installed with CUDA 12.1, cuSPARSELt 0.4.0, PyTorch 2.1.0, Transformers v4.40.0 and vLLM 0.4.0.post1. 
The GPU used in the evaluation is NVIDIA GeForce RTX 4070 Super (except in \S\ref{sec:HardwarePortability}).
The CPU frequency scaling is disabled in all experiments for fairness.

\textbf{Baselines.} \textit{Kernel level:} The baselines for kernel level performance evaluation consist of several SOTA dense and sparse kernel libraries, including \textit{cuBLAS}\cite{cublas}, \textit{Sputnik}\cite{Sputnik}, \textit{cuSPARSELt}\cite{cusparselt} and \textit{VENOM}\cite{VENOM}. \textit{cuBLAS} and \textit{cuSPARSELt} are black-box vendor-specific libraries provided by NVIDIA, which are manually written and optimized with expertise and present the peak performance on NVIDIA GPUs for dense and structured sparse operations, respectively. \textit{Sputnik} is a leading open source library targeting accelerating the sparse operations in deep learning applications maintained by Google. \textit{VENOM}, proposed in late 2023, accelerates sparse matrix multiplication with SpTC hardware, providing 1.38$\times$ speedup compared to the cuSPARSELt library.

\textit{Model level:} We compare Samoyeds against three leading solutions: \textit{Transformers}\cite{Transformers}, the most popular framework for computing LLM models, with the latest released version (v4.40.0, on April 18, 2024); \textit{MegaBlocks}\cite{megablocks}, which is specifically designed and optimized for block-sparse operations in MoE computations, surpassing solutions like Tutel\cite{hwang2023tutel} and Megaron-LM\cite{megatronlm}; and \textit{vLLM-DS}\cite{vllm, dai2024deepseekmoe} which provides a SOTA fused kernel for the MoE process. It should be noticed that baseline vLLM-DS integrates the recent implementation and optimization of a fused kernel for MoE models (merged on \textit{March 2, 2024}), achieving approximately a 2.8$\times$ speedup compared with previous non-fusion version\cite{vllm-ds}. To ensure fairness, all experiments in model level employs Flash-Attention2 in the decoder layer.

\begin{table}%\scriptsize
    \caption{Configurations of MoE Models.}
    \label{tab:kernel.config}
    \resizebox{\linewidth}{!}{
        \begin{tabular}{ccccc}
            \toprule
            Model & Expert & Hidden Size & Intermediate Size & Config Num\\
            \midrule
            Qwen2-MoE           & 60    & 1408  & 2048  & \multirow{2}{*}{CFG\#1} \\
            DeepSeek-MoE        & 64    & 1408  & 2048  & \\
            MiniCPM-MoE         & 8     & 2304  & 5760  & CFG\#2\\
            OpenMoE-34B         & 32    & 3072  & 12288 & CFG\#3\\
            Mixtral-8$\times$7B & 8     & 4096  & 14336 & CFG\#4\\
            Mixtral-8$\times$22B& 8     & 6144  & 16384 & CFG\#5\\
          \bottomrule
        \end{tabular}
    }
\end{table}

\subsection{Kernel Performance} \label{sec:kernel}

\begin{figure}
    \centering
    \includegraphics[width=0.98\linewidth]{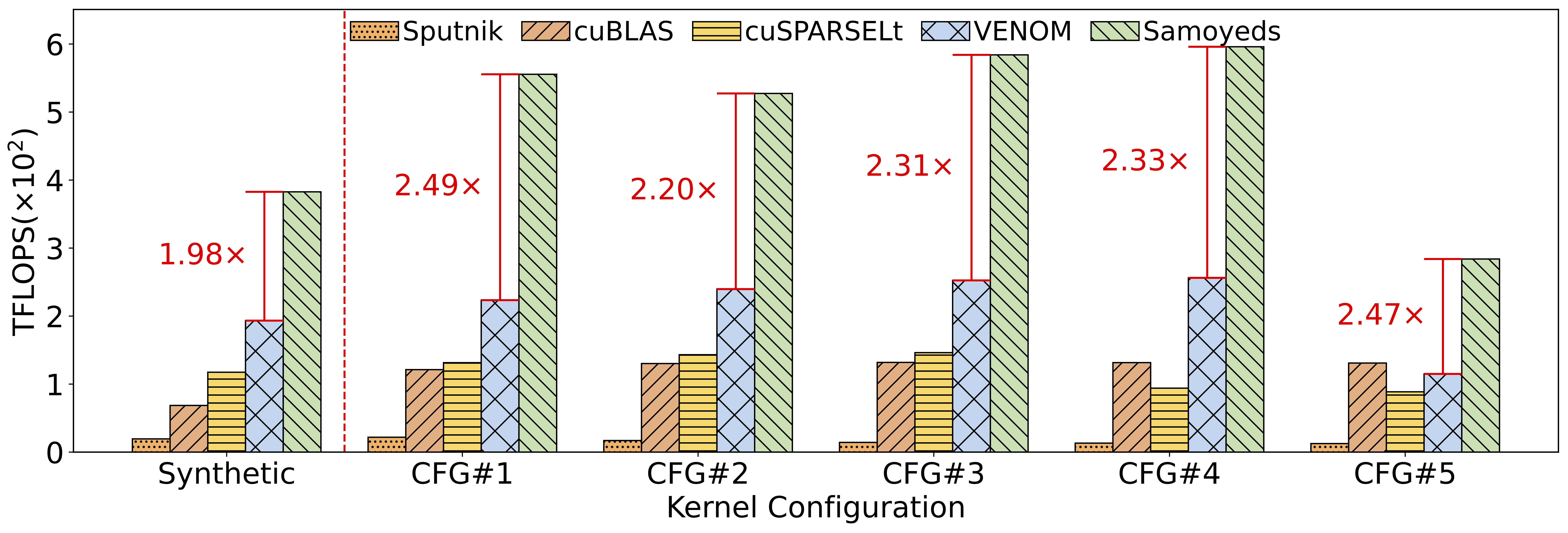}
    \caption{Kernel Performance Comparison on Synthetic and Realistic Benchmarks. \textnormal{The synthetic benchmark covers 238 sizes; the realistic benchmark reflects typical model sizes.}}
    \label{fig:exp.kernel}
\end{figure}

\subsubsection{Overall Kernel Performance} \label{sec:kernel_overall} 

The overall kernel-level performance of Samoyeds is evaluated on the synthetic benchmark to reveal the benefits of Samoyeds in wide application scenarios, and the realistic benchmark to present the performance benefits when applied to specific models. 
The synthetic benchmark consists of 238 distinct cases, with dimensions $m$, $k$, and $n$ ranging from 256 to 16384.
The cases in the realistic benchmark are extracted from MoE LLMs as detailed in Table \ref{tab:kernel.config}.

The results are illustrated in Figure \ref{fig:exp.kernel}, demonstrating that Samoyeds consistently outperforms other baselines. For synthetic benchmark, shown on the left of Figure \ref{fig:exp.kernel}, compared to the SOTA sparse matrix library VENOM, Samoyeds kernel exhibits a speedup of up to 1.99$\times$. Furthermore, the relative speedup over cuBLAS, cuSPARSELt, and Sputnik reaches as high as 5.44$\times$, 3.18$\times$, and 18.76$\times$, respectively. The results on the realistic benchmark are shown on the right of Figure \ref{fig:exp.kernel}. Samoyeds provides an average speedup of 2.33$\times$ (peaking at 2.49$\times$) compared to the best baseline, VENOM. Compared to kernel libraries cuBLAS and cuSPARSELt, Samoyeds achieves average speedups of 3.95$\times$ and 4.29$\times$, respectively. Despite being specifically optimized for sparse operations in deep learning, Sputnik still shows poor performance because it fails to leverage the structured pattern of the sparse data and hardware features. In contrast, Samoyeds significantly outperforms Sputnik, delivering an average speedup of 33.02$\times$.

It is noteworthy that, under CFG\#5, the overall throughput of both Samoyeds and VENOM is lower than that observed with other configurations. This performance degradation is mainly caused by the skewness between dimensions $m$ and $n$, with $m$ being substantially larger. This imbalance leads to decreased memory efficiency during computation, a corner case that could be mitigated with a more sophisticated tiling implementation. It also adversely affects the throughput of other baseline kernels. Despite this, Samoyeds still offers a significant speed advantage over cuBLAS and cuSPARSELt under CFG\#5, where VENOM does not, achieving a speedup of 2.47$\times$ relative to VENOM. These findings underscore the efficiency and practical applicability of Samoyeds in real-world computing scenarios, demonstrating its superiority over existing approaches in handling MoE computations.

\begin{figure}[t]
    \centering
    \includegraphics[width=\linewidth]{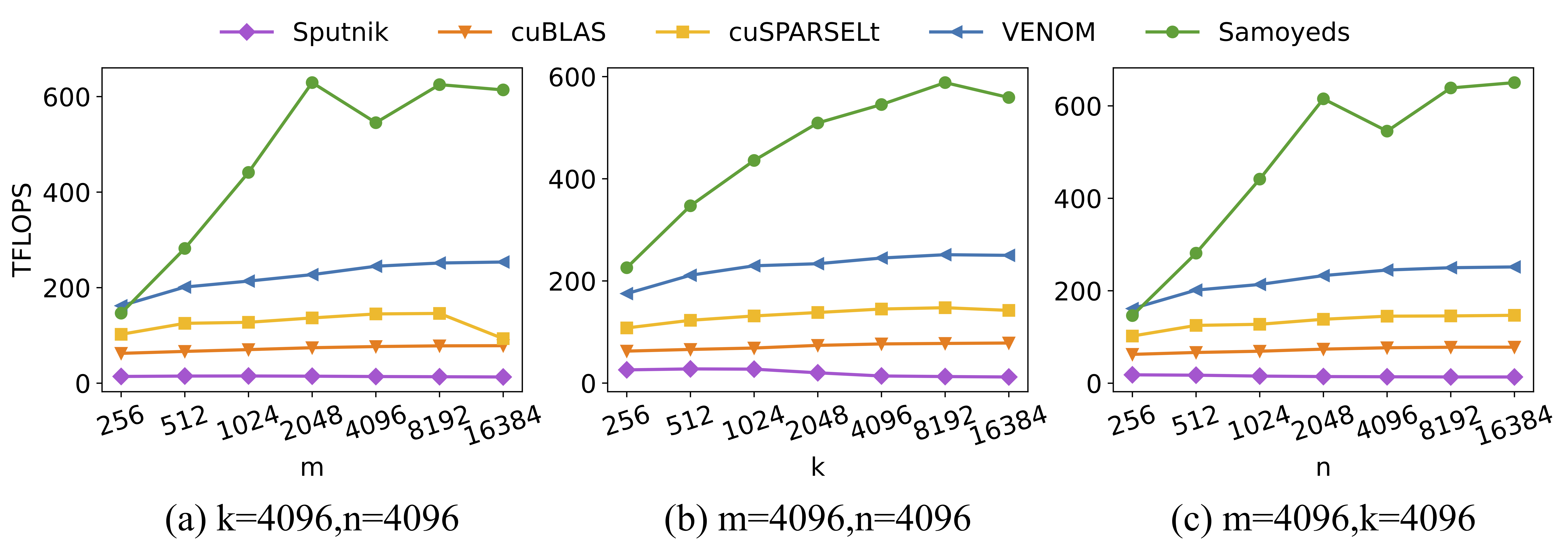}
    \caption{Throughput Trend with Varying Operator Size.}
    \label{fig:exp.kernel.trend} 
\end{figure}

\subsubsection{Throughput with Different Operator Sizes} \label{sec:kernel_trend}

We investigate the performance trends of Samoyeds kernel by profiling the throughput as matrix dimensions increase. The results, as illustrated in Figure \ref{fig:exp.kernel.trend}, demonstrate that Samoyeds consistently outperforms all other baselines across nearly all matrix sizes. When compared to the best baseline, VENOM, Samoyeds achieves a speedup of up to 2.77$\times$ within a varying range of dimension $m$, and up to 2.34$\times$ and 2.58$\times$ for dimensions $k$ and $n$, respectively.

As operand sizes scale up, the throughput of Samoyeds kernel increases significantly before reaching the peak performance. The reason for this increment varies with dimension. For dimension $k$, the initialization and write-back overhead of the kernel remains constant, allowing these costs to be amortized over a growing number of reduction computations. Consequently, the throughput of Samoyeds asymptotically approaches its maximum value as shown in Figure \ref{fig:exp.kernel.trend}(b). For dimensions $m$ and $n$, the number of warps per kernel increases as the operand sizes get larger. This increase allows the hardware warp scheduler to select from a larger pool of active warps when the current warp stalls, thereby optimizing resource utilization. This is reflected in Figure \ref{fig:exp.kernel.trend}(a) \& \ref{fig:exp.kernel.trend}(c), where throughput scales linearly with dimensions $m$ and $n$ before it reaches a peak, benefiting from increased parallelism. These trends highlight the efficiency and scalability of Samoyeds across a wide range of matrix dimensions, underscoring its robust applicability in scenarios that require high computational throughput.

Notably, when dimensions $m$ or $n$ are set to 256, Samoyeds slightly underperforms compared to VENOM due to the limited parallelism available at this kernel size, which can be addressed by implementing smaller kernel tiling sizes. Additionally, unlike other baselines presenting stable peak performance across varying operand sizes, fluctuations occur in Samoyeds kernels because our kernel has not been specially adapted for corner cases, indicating the potential for further refinement in Samoyeds kernels. 
Moreover, as shown in Figures \ref{fig:exp.kernel.trend}(a) and \ref{fig:exp.kernel.trend}(c), the performance exhibits a slight decline when the corresponding dimension size is 4096. This behavior can be explained by two factors: (1) as the size increases from 2048 to 4096, scheduling more warps on each SM leads to L1 cache eviction when switching between warps, which significantly reducing the cache hit rate by 76.38\%, thereby degrading overall performance; (2) when the size further increases to 8192, the larger problem size enables better scheduling opportunities and amortizes the tail wave overhead, resulting in a 5.90\% increase in active warps per scheduler and improved performance.

\subsection{MoE Layer Performance} \label{sec:MoELayer}

\begin{figure*}[t]
    \centering
    \includegraphics[width=0.98\linewidth]{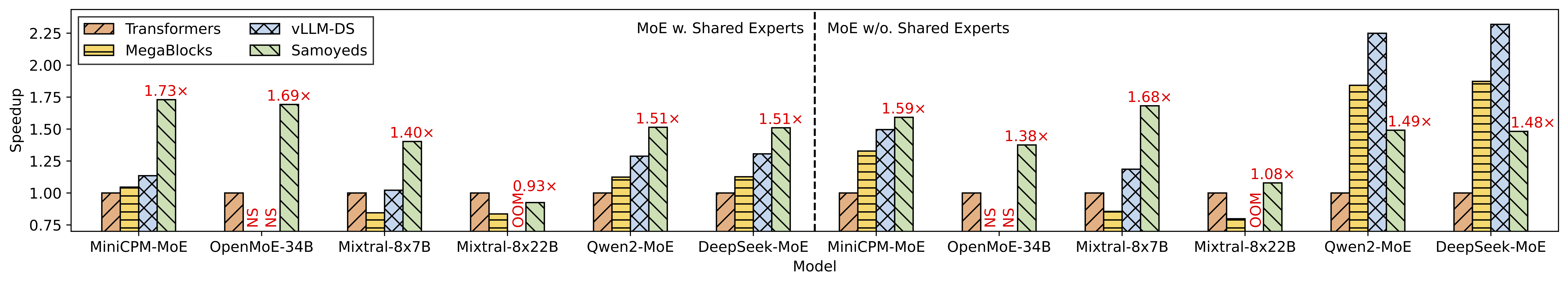} 
    \caption{Execution Speedup for the MoE Layer. \textnormal{\textit{NS} indicates \textit{not supported} due to kernel incompatibility with OpenMoE-34B. \textit{OOM} denotes out-of-memory errors preventing execution completion.}}
    \label{fig:exp.MoE} 
\end{figure*}

The current SOTA LLMs that employ the MoE technique can be categorized into two types. The first type utilizes multiple experts with identical routing priorities, where each token is dispatched to specific experts based on routing outcomes. The second type incorporates several isolated shared experts alongside individually routed experts. This setup enforces all tokens to be processed by all these shared experts in addition to their assigned routed experts. Therefore, it is crucial to demonstrate the effectiveness of Samoyeds under these two distinct routing types.

Figure \ref{fig:exp.MoE} shows the speedup achieved on the MoE layer compared to the original Transformers solution. The configuration of the size and number of routed experts aligns with model configurations in Table \ref{tab:kernel.config}, with the number of tokens set as 4096. The left of the figure displays results for MoE layers incorporating two isolated shared experts, while the right shows results for MoE layers without shared experts. 

With shared experts enabled, Samoyeds achieves an average speedup of 1.46$\times$, peaking at 1.73$\times$ compared to the most widely-used Transformers. Additionally, it outperforms MegaBlocks and the SOTA baseline vLLM-DS, providing a speedup of up to 1.66$\times$ and 1.53$\times$, respectively. Without shared experts, Samoyeds achieves an average speedup of 1.45$\times$ and peaks at 1.68$\times$ compared to Transformers. Furthermore, Samoyeds surpasses other baselines in most models, including vLLM-DS.

Notably, the \textit{NS} marker denotes \textit{Not Supported}, as the OpenMoE-34B features a distinct activation function within the MLP layer, which is incompatible with the specialized kernels provided by MegaBlocks and vLLM-DS. Additionally, the OOM (Out-of-Memory) marker indicates that the corresponding solution failed to complete the computing process due to memory constraints. Samoyeds underperforms slightly when applied in the Mixtral-8$\times$22B, due to the skewness in expert size, a tiling corner case previously discussed in \S\ref{sec:kernel_overall}. This issue could be mitigated by further implementing different tiling sizes. In the same configuration, MegaBlocks suffers from significant slowdowns, and vLLM-DS encounters OOM errors, which highlights the robustness of Samoyeds in these conditions. 

Moreover, in Qwen2-MoE and DeepSeek-MoE without shared experts, the speedup advantage of Samoyeds is less evident when compared to MegaBlocks and vLLM. This reduced performance can be attributed to two factors. Firstly, the smaller size of experts in these models results in lower parallelism, as discussed in \S\ref{sec:kernel_trend}, leading to reduced computation speeds. Secondly, the number of tokens processed by each expert must be aligned with the tiling size of kernels. If this number does not align perfectly, the original input must be supplemented with empty tokens, creating extra computational work. Therefore, models with more experts will suffer severe padding overhead, considering the number of tokens each expert needs to process decreases. By simply implementing a smaller tiling size, the extra padding overhead can be saved, indicating further potential speedup improvements in Samoyeds.

\begin{figure}[t]
    \includegraphics[width=0.98\linewidth]{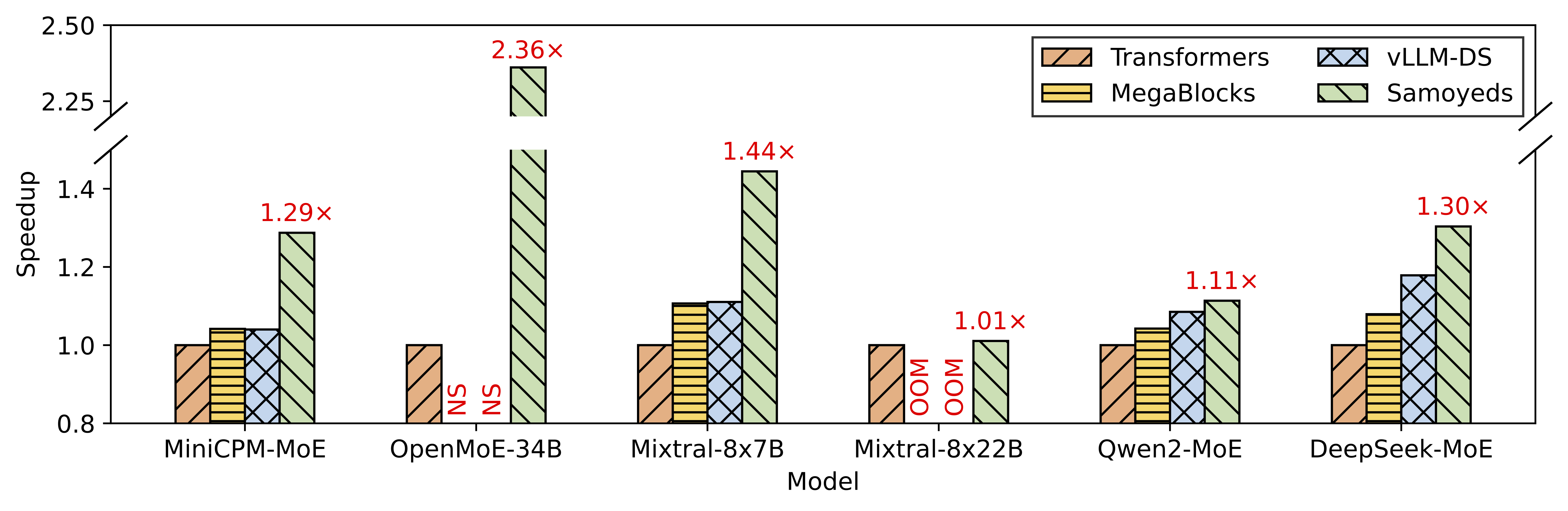} 
    \setlength{\abovecaptionskip}{0.1cm}
    \caption{Speedup in End-to-end Latency of MoE Models.}
    \label{fig:exp.layer.latency}
\end{figure}
  
\subsection{End-to-End Model Evaluation} \label{sec:E2E}

We benchmark the end-to-end performance of Samoyeds and the baselines on various real-world leading MoE LLMs shown in Table \ref{tab:kernel.config}. 
To accommodate the memory capacity constraints of GPUs, we measure the performance of a single decoder layer. 
This is justified by two observations: (1) prevalent MoE LLMs are decoder-only, with decoder layers accounting for over 90\% of the total execution time and (2) the decoder layers share similar architectures and sizes, leading to comparable execution times.
Notably, the input for Samoyeds and other baselines remains consistent, ensuring identical routing distributions and guaranteeing a fair comparison.

\subsubsection{Overall Model Performance.} \label{sec:E2E_overall}

We initially compare the overall performance of these models using Samoyeds and other baselines, with a default sequence length of 4096 and a batch size of 1. For the OpenMoE-34B, we adjust the sequence length to 2048 due to its maximum sequence length constraints. Additionally, for DeepSeek-MoE and Qwen2-MoE, we increase the batch size to 16 to leverage the larger number of experts within these models. MegaBlocks and vLLM-DS are not supported in OpenMoE-34B due to incompatibility. Meanwhile, they both fail to complete processing Mixtral-8$\times$22B due to OOM errors.

As illustrated in Figure \ref{fig:exp.layer.latency}, Samoyeds significantly outperforms all competing baselines. In particular, Samoyeds achieves a remarkable speedup of up to 2.36$\times$ (1.42$\times$ on average) compared to Transformers. Additionally, it delivers speedup of up to 1.31$\times$ and 1.30$\times$ relative to MegaBlocks and the SOTA baseline vLLM-DS, respectively. These results highlight the effectiveness of our optimization strategies in enhancing performance.

\begin{figure}[t]
    \includegraphics[width=0.98\linewidth]{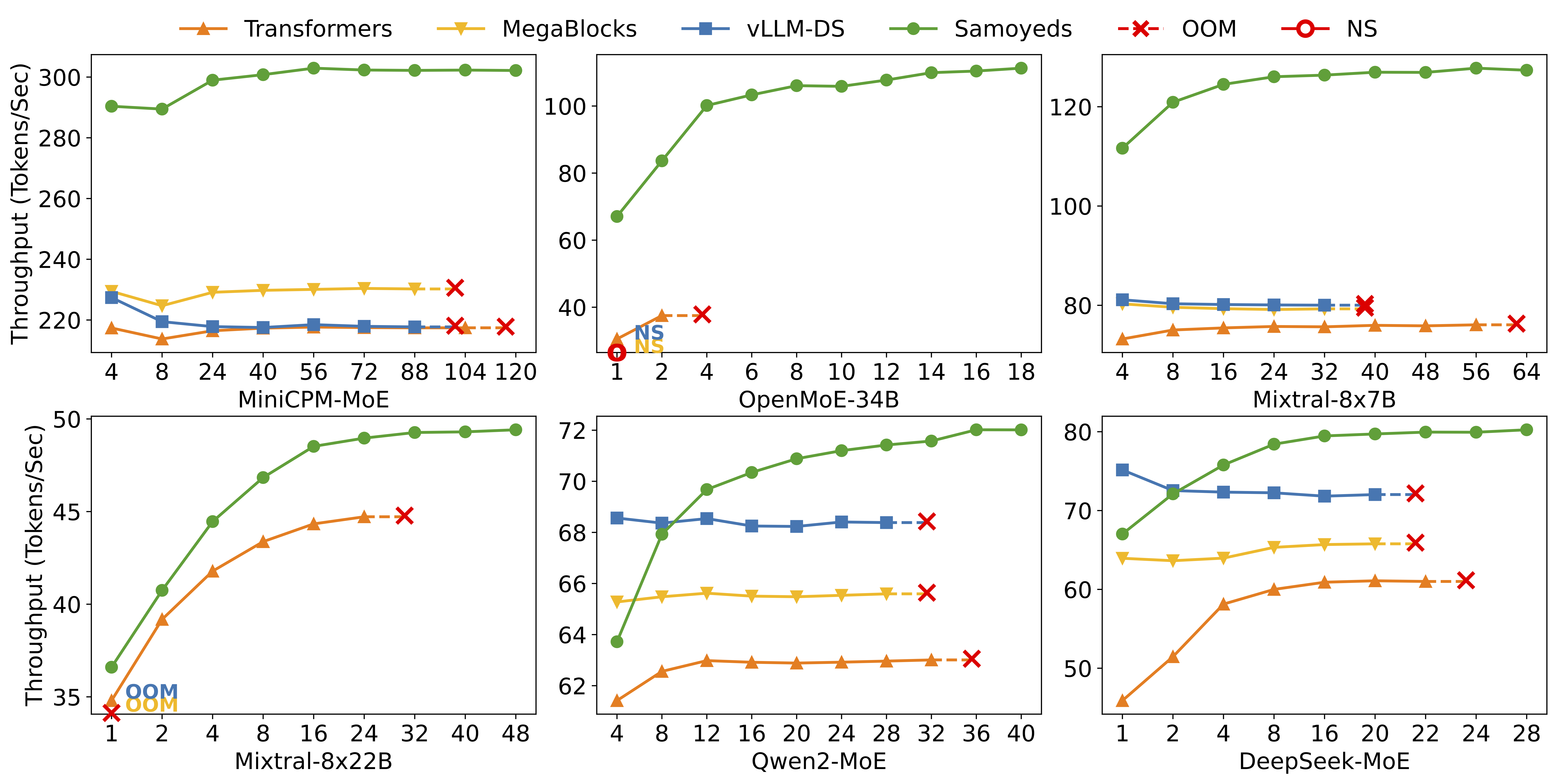} 
    \caption{Throughput under Different Batch Sizes.}
    \label{fig:exp.layer.batch}
\end{figure}

\begin{table} %\scriptsize
  \caption{Maximum Batch Sizes for MoE Models.}
  \label{tab:model.batch.size}
  \resizebox{\linewidth}{!}{
      \begin{tabular}{cccccc}
        \toprule
        Model & Transformers & MegaBlocks & vLLM-DS & Samoyeds & \makecell[c]{Boost over the \\Best Baseline}\\
        \midrule
        MiniCPM-MoE         & 118   & 89    & 91    & \textbf{123}  & 1.04$\times$\\
        OpenMoE-34B         & 3     & -     & -     & \textbf{56}   & 18.67$\times$ \\
        Mixtral-8$\times$7B & 62    & 36    & 36    & \textbf{86}   & 1.38$\times$ \\
        Mixtral-8$\times$22B& 30    & 0     & 0     & \textbf{53}   & 1.77$\times$ \\
        Qwen2-MoE           & 35    & 28    & 28    & \textbf{44}   & 1.26$\times$ \\
        DeepSeek-MoE        & 22    & 21    & 21    & \textbf{52}   & 2.36$\times$ \\
      \bottomrule
      \end{tabular}
  }
\end{table}

\subsubsection{Throughput with Different Batch Sizes.} \label{sec:E2E_batch_size}
We explore the throughput of various models across different batch sizes, as illustrated in Figure \ref{fig:exp.layer.batch}. For models equipped with smaller expert configurations, including Qwen2-MoE and DeepSeek-MoE, we maintain a sequence length of 4096 per batch. Conversely, for other models featuring larger experts, we reduce the sequence length to 1024 per batch to provide a clearer insight into throughput trends with increasing batch sizes. 
OpenMoE-34B is not supported by MegaBlocks and vLLM due to incompatibility.

Our method, Samoyeds, shows superior throughput compared to other baselines across a variety of configurations and batch sizes. Specifically, Samoyeds achieves significant speedups over the best baseline in all models as batch size increases. The speedup of different models is up to 1.31$\times$, 2.23$\times$, 1.58$\times$, 1.09$\times$, 1.04$\times$, and 1.11$\times$, compared to the best baseline, respectively. Notably, the throughput using MegaBlocks and vLLM-DS shows minimal fluctuation along with batch size increasing, in contrast, the throughput using Samoyeds method increases significantly before reaching a stable peak. The underlying reason for these observations is the improved parallelism as discussed previously in \S\ref{sec:kernel_trend}.

Furthermore, as illustrated in Table \ref{tab:model.batch.size}, the maximum batch size supported by Samoyeds exceeds that of other methods. Compared to Transformers, Samoyeds supports a significantly wider range of batch sizes (4.41$\times$ larger on average). Interestingly, the boost in maximum batch size of OpenMoE-34B is exceptionally higher, likely due to its unique computation process compared to other models. Notably, although approaches like MegaBlocks and vLLM-DS can accelerate model execution over Transformers, the maximum batch size supported for these approaches significantly decreases. They even fail to complete computations for Mixtral-8$\times$22B with the batch size set to 1. This finding highlights the superior efficiency of Samoyeds in memory utilization, which in turn enhances its ability to process more batches concurrently.

% TODO 对应说法修改一下
\begin{figure}[t]
  \includegraphics[width=\linewidth]{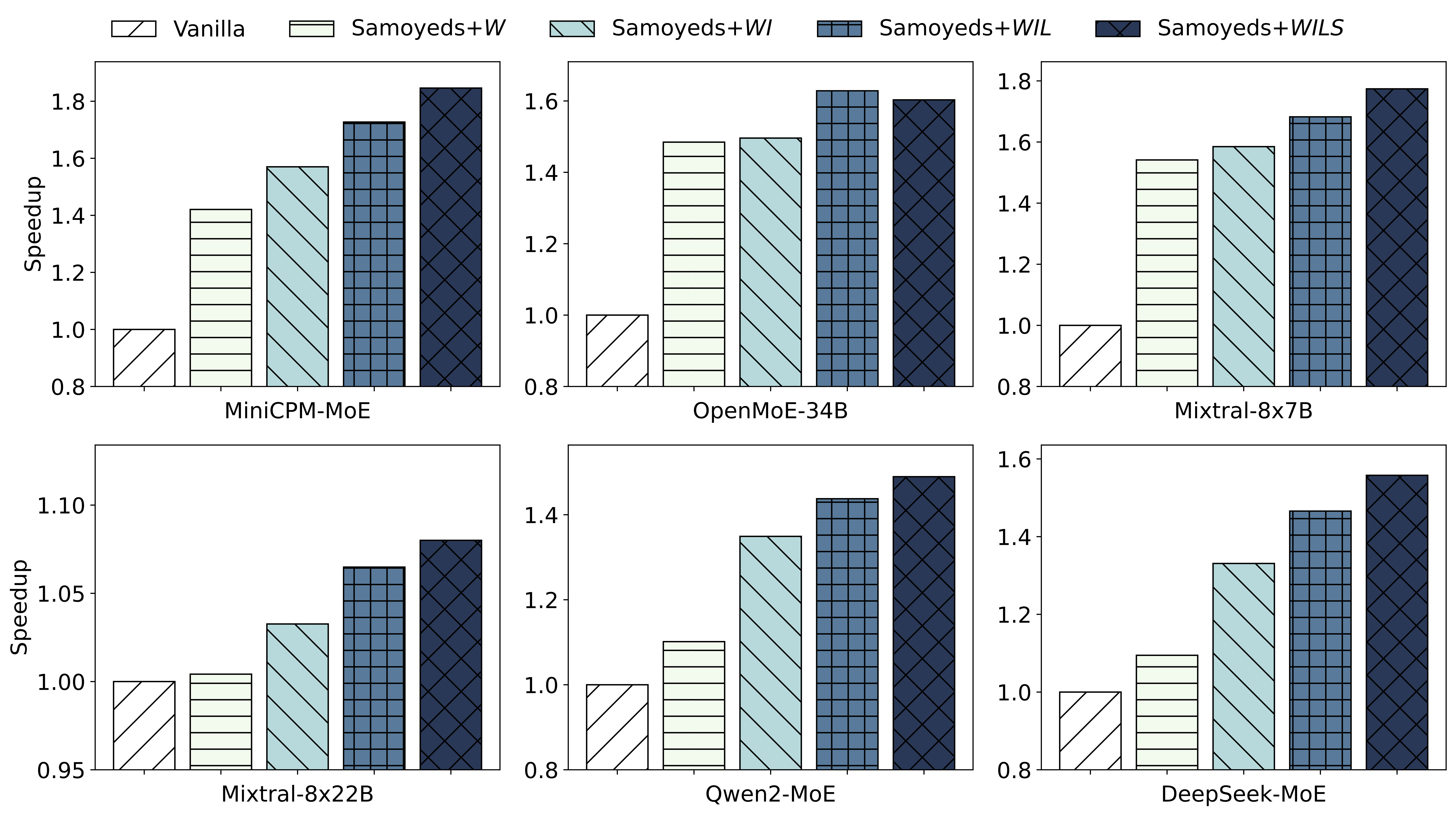}
  \caption{Breakdown Analysis on Samoyeds Optimizations. Methods are denoted by abbreviation letters. \textit{W}: weight sparsity, \textit{I}: input sparsity, \textit{L}: data layout, \textit{S}: data stationary.}
  \label{fig:exp.kernel.ablation}
\end{figure}

\subsection{Breakdown Analysis} \label{sec:ablation}

In this section, we break down the performance enhancements brought by Samoyeds. \textit{Vanilla} represents the standard Transformers framework. Then the optimizations are enabled step by step as illustrated in Figure \ref{fig:exp.kernel.ablation}. 

We first enable leveraging sparsity in model weights, denoted as \textit{Samoyeds+$W$}, by utilizing the kernel for sparse-dense matrix multiplication. The introduction of weight pruning (\textit{Samoyeds+$W$}), discussed in \S\ref{sec:MoE_sparse_weight}, results in an average speedup of 1.27$\times$ over \textit{Vanilla}, peaking at 1.54$\times$.

Next, we eliminate the redundancy in data flow, labeled as \textit{Samoyeds+$WI$}, by adopting the sparse-sparse matrix multiplication kernel. By eliminating the input permutation overhead, discussed in \S\ref{sec:MoE_sparse_input}, \textit{Samoyeds+$WI$} enhances performance by 1.39$\times$ on average compared to the \textit{Vanilla} method. This configuration also outperforms \textit{Samoyeds+$W$} in all tested models, with speedups reaching up to 1.23$\times$. Notably, models with more experts, such as Qwen2-MoE and DeepSeek-MoE, experience a greater performance benefit due to their amplified performance loss from input permutation.

Furthermore, we evaluate the benefits of reducing transposition overhead, denoted as \textit{Samoyeds+$WIT$}. With this graph-level optimization as previously discussed in \S\ref{sec:data_layout_opti}, \textit{Samoyeds+$WIT$} improves performance by up to 1.08$\times$ on average compared to \textit{Samoyeds+$WI$}.

Finally, we incorporate the data stationary optimization referred to as \textit{Samoyeds+$WITS$}. Overall, the increased data reuse, as discussed in \S\ref{sec:data_stationary}, delivers an average speedup of 1.04$\times$ over the \textit{Vanilla} approach. 

\subsection{Accuracy Assessment}\label{sec:accuracy}

In this section, we first prune the model using the proposed Samoyeds sparse format to evaluate model accuracy. The inference solutions proposed in Samoyeds are fully decoupled from the pruning process, enabling seamless integration with SOTA pruning method such as WoodFisher\cite{NEURIPS2020_d1ff1ec8}, which is based on second-order pruning and SparseGPT\cite{FrantarA23}, which operates without gradient information. 
In our experiments, we use the WoodFisher method provided by the SparseML framework. Notably, WoodFisher incurs significantly higher memory usage during pruning compared to other methods.
Therefore, we select the most representative models within the models that are feasible under a limited memory budget, including Bert, Tiny-LLaMA and Qwen2-1.5B. Moreover, as demonstrated in prior research\cite{mishra2021accelerating, NEURIPS2020_d1ff1ec8, FrantarA23}, maintaining accuracy during pruning is more challenging for smaller models, making them a compelling choice for evaluating the effectiveness of our approach.
To ensure fairness, a uniform sparsity ratio of 75\% is enforced across all methods, excluding the dense baseline.

First, we analyze model accuracy across different sparse configurations. The sparse format is denoted as (N,M,V), corresponding to the structured sparse granularity configuration introduced in Section \ref{sec:sparse_encoding}. As shown in Table \ref{tab:acc.bert}, the accuracy of BERT models remains stable under varying sparse configurations. On the SQuAD 1.1 task, the Samoyeds sparse format retains over 99.3\% of the original accuracy on average. Additionally, the accuracy of models pruned with Samoyeds sparse format is comparable to that of dense models and those pruned with unstructured methods (magnitude-based)\cite{hagiwara1994simple, NIPS2015_ae0eb3ee}. As shown in Table \ref{tab:acc.ppl}, the increase in perplexity for the  GSM8K text generation tasks is only 0.06 and 0.05 for the Tiny-LLaMA-1B and Qwen2-1.5B models, respectively. Notably, models pruned with the Samoyeds sparse format outperform those pruned with the SOTA structured pruning method, VENOM\cite{VENOM}, by 56\% and 73\%, respectively.

\begin{table}\scriptsize
  \caption{F1 Score of Bert-Series Models pruned with different Samoyeds configurations on SQuAD 1.1 (higher is better).}
  \label{tab:acc.bert}
  \resizebox{\linewidth}{!}{
      \begin{tabular}{cccccc}
        \toprule
        Model & Dense & (1,2,16) & (1,2,32) & (4,8,32) & (8,16,32)  \\
        \midrule
        Bert-base   & 89.50 & \textbf{88.83} & 88.48 & 88.57 & 88.60 \\
        Bert-large  & 88.86 & 88.26 & \textbf{88.66} & 88.25 & 88.51 \\
      \bottomrule
      \end{tabular}
  }
\end{table}

\begin{table}\scriptsize
  \caption{Perplexity of Models pruned into different formats on GSM8K (lower is better).}
  \label{tab:acc.ppl}
  \resizebox{\linewidth}{!}{
      \begin{tabular}{cccccc}
        \toprule
        Model & Dense & Unstructured & VENOM &  Samoyeds  \\
        \midrule
        Tiny-LLaMA  &  1.72 & 1.94  & 1.95  & 1.82 \\
        Qwen2       &  1.92 & 1.96  & 2.26  & 2.01 \\
      \bottomrule
      \end{tabular}
  }
\end{table}

\begin{figure}[t]
    \centering
    \begin{minipage}[b]{0.49\linewidth}
        \centering
        \setlength{\abovecaptionskip}{0.2cm}
        \includegraphics[width=\linewidth]{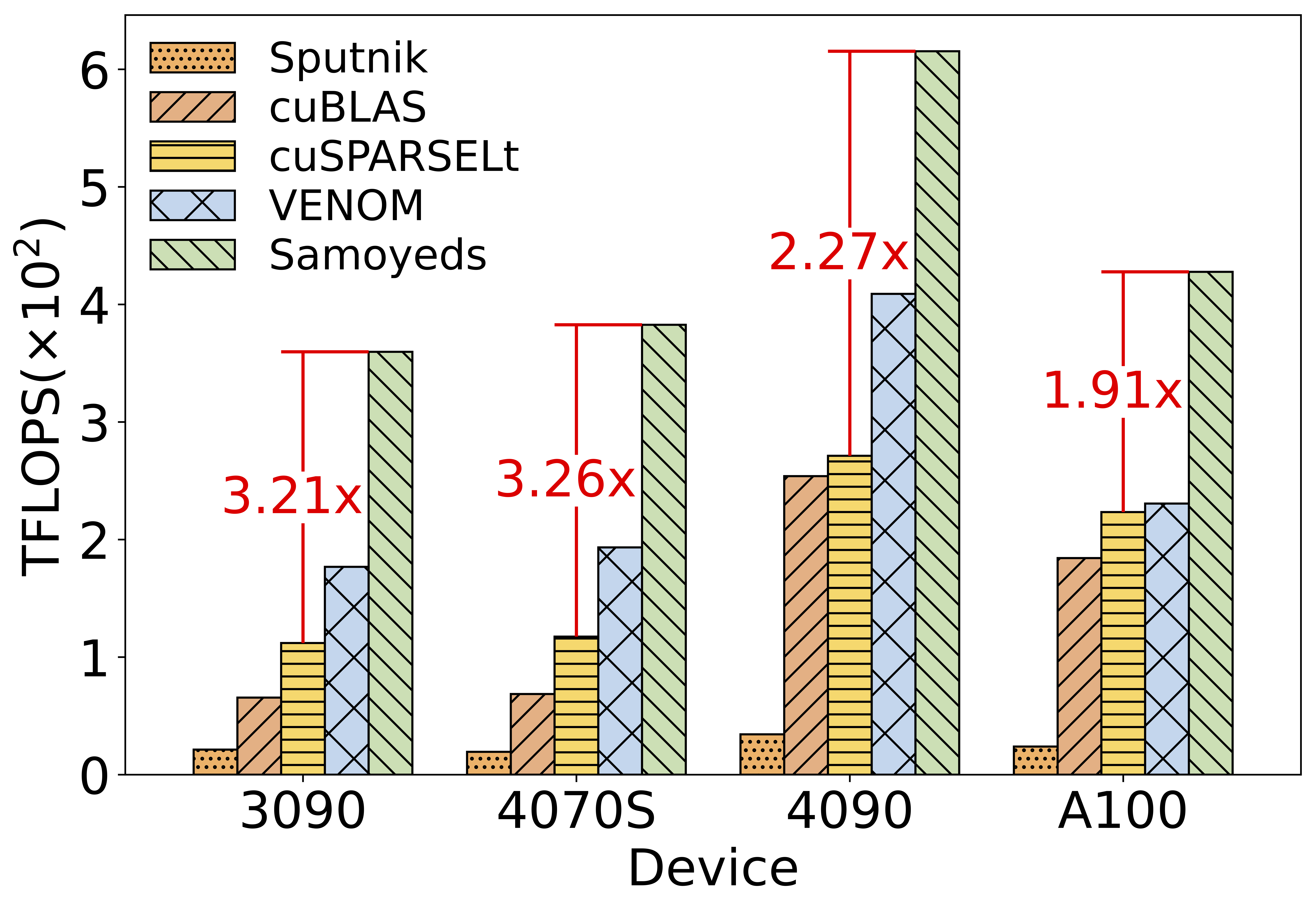} 
        \caption{Performance with Direct Porting.}
        \label{fig:exp.kernel.devices} 
    \end{minipage}
    \hfill
    \begin{minipage}[b]{0.49\linewidth}
        \centering
        \setlength{\abovecaptionskip}{0.2cm}
        \includegraphics[width=\linewidth]{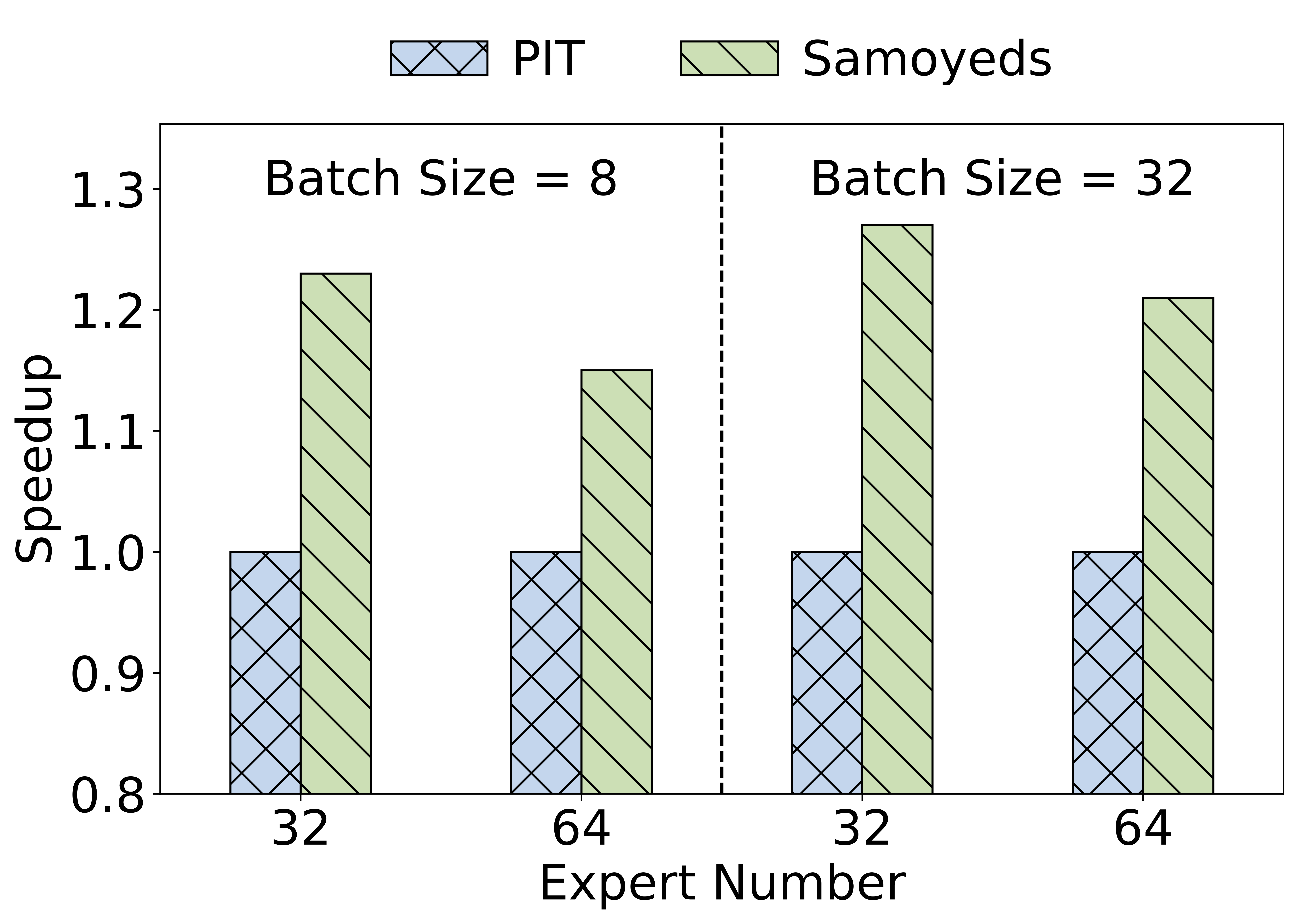} 
        \caption{Performance Comparison with PIT.}
        \label{fig:exp.PIT}
    \end{minipage}
\end{figure}

\begin{table}\scriptsize
\caption{Performance Portability under Suggested Adaptations. The results show the percentage of the synthetic set with improved, unchanged, or degraded performance after applying the adaptation.}
\label{tab:arch.change}
\centering
\resizebox{\linewidth}{!}{
\begin{tabular}{cccccc}
\toprule 
\multirow{2}{*}{\makecell{Porting \\Target}} & \multirow{2}{*}{\makecell{Hardware \\ Differences}} & \multirow{2}{*}{Adaptation} & 
\multicolumn{3}{c}{Perf. Impact on Cases}  \\
& & & Improved  & Unchanged & Degraded \\
\midrule
\textbf{A100}   & \makecell[c]{SM $\uparrow$ \\L2 Cache $\downarrow$} & Tile Size $\downarrow$ & 55.9\% & 5.5\% & 38.6\% \\
\midrule
\textbf{3090}   & \makecell[c]{Slower TC \\Bandwidth $\uparrow$}   & Stage Num $\uparrow$ & 39.1\% & 49.6\% & 11.3\% \\
\bottomrule
\end{tabular}
}
\end{table}

\subsection{Performance Portability Analysis}\label{sec:HardwarePortability}
In this section, we assess the performance portability of Samoyeds across various hardware platforms with similar micro-architectures, including NVIDIA 3090, 4070 Super, 4090 and A100 40G GPUs. We directly port the kernel implementation on 4070S to other hardware and evaluate the performance using the synthetic dataset from \S\ref{sec:kernel_overall}, which contains 238 distinct problem sizes. As shown in Figure \ref{fig:exp.kernel.devices}, Samoyeds maintains 65.2\% of its relative speedup over cuSPARSELt on average, with 41.0\% retained in the worst-case scenario. In contrast, VENOM loses 95\% relative speedup on A100, exhibiting almost no improvement compared to cuSPARSELt. 
This performance discrepancy stems from two key factors: (1) While vendor libraries (e.g., cuBLAS, cuSPARSELt) employ hardware-specific kernel configurations across different GPUs, both VENOM and Samoyeds are primarily optimized for their native development platforms. This architectural specialization inevitably diminishes their performance gains when deployed on different hardware. (2) VENOM suffers from memory-computation imbalance when porting to A100 as this GPU is equipped with higher memory bandwidth but slower tensor cores, which increases pipeline stalls during execution. However, Samoyeds mitigates this imbalance through its sparse memory access paradigm, leading to better portability when porting to A100.

Additionally, we further explore the potential adaptation rules to improve the performance given different hardware configurations.
Specifically, the tiling size hyper-parameter affects the utilization of streaming multiprocessors (SMs) in parallel and the L2 cache hit rate. Meanwhile, the tensor core processing speed and memory bandwidth can affect the overlapping of the pipeline stages. 
As illustrated in Table \ref{tab:arch.change}, we propose several suggested adaptations for porting to different devices and evaluate the performance with and without these adaptations using the synthetic set described in \S \ref{sec:kernel_overall}. For instance, A100 GPU features more SMs but has a smaller L2 cache compared to 4070S. To fully exploit the parallelism of A100 and improve the L2 cache hit rate, it is suggested to employ smaller tiling sizes, which can lead to a performance boost in 55.9\% of the tested cases.

\subsection{Comparison with Compiler for Dynamic Sparsity}\label{sec:ComparisonwithPIT}

In this section, we compare the performance of Samoyeds against the SOTA compiler-based solution, PIT\cite{PIT}, which is specifically designed to leverage the dynamic sparse pattern that emerges in the execution of LLMs. It aggregates multiple sparse micro-tiles into dense tiles with its \textit{Permutation Invariant Transformation}, improving overall GPU utilization. Figure \ref{fig:exp.PIT} illustrates the normalized speedup of the MoE layer with different batch sizes and expert numbers. Samoyeds outperforms PIT by 1.15$\times$ to 1.27$\times$, depending on the configuration.

It should be noted that while PIT claims theoretical support leveraging the sparsity pattern along three dimensions for matrix multiplication, its practical implementation is limited to permutation along one dimension. Furthermore, PIT does not integrate the SpTC hardware into its compiler to further leverage the sparse computing capability of hardware. Consequently, PIT can only exploit the sparsity pattern that dynamically emerges in activations, which makes Samoyeds naturally outperform PIT, as demonstrated in our evaluation.

\section{Related Work}

\textbf{Leveraging sparsity in LLMs.} 
To reduce computation costs in LLMs, recent works focus on leveraging sparse computation. For unstructured sparsity, cuSPARSE\cite{cusparse} and Sputnik\cite{Sputnik} provide GPU-accelerated linear algebra subroutines optimized for deep learning. For structured sparsity, some studies \cite{mao2017exploring, narang2017exploring, ChenQLDX21, SparTA} focus on leveraging sparsity in model parameters, while others\cite{PIT, dejavu, ProSparse} explore the sparsity that dynamically emerges in activations during model execution.
The N:M structured sparsity has gained attention for its benefits in boosting computation efficiency while preserving model accuracy\cite{mishra2021accelerating}. Libraries such as DFSS\cite{ChenQQ0DX23}, nmSPARSE\cite{nmSparse} and cuSPARSELt\cite{cusparselt} optimize kernels leveraging SpTC hardware with 2:4 sparsity. Besides, VENOM\cite{VENOM} extends the capabilities of SpTC by supporting flexible sparse ratios. 

\textbf{MoE optimizations.} 
Prior works on MoE optimizations fall into three categories. The first involves designing routing algorithms to enhance model accuracy, such as employing hash layers\cite{Hashmoe}, random routing\cite{thor} or reinforcement learning\cite{ClarkCGMPHDHCB022}. The second focuses on enhancing parallelism mechanisms, with approaches like addressing communication overhead in Lina\cite{li2023accelerating}, employing dynamic shadowing in FasterMoE\cite{he2022fastermoe} and automatically discovering optimal parallel strategies in SmartMoE\cite{zhai2023smartmoe}. The third optimizes MoE computation directly, like the introduction of block-sparse operations in MegaBlocks\cite{megablocks} and the design of a faster kernel for MoE layers in DeepseekMoE\cite{dai2024deepseekmoe}. 

\section{Conclusion}
This paper presents Samoyeds, a novel acceleration system for MoE LLMs with software-hardware co-optimization. 
We introduce a new sparse format tailored to the dual-sided sparsity inherent in MoE LLMs and implement a bespoke sparse-sparse multiplication kernel leveraging SpTC to eliminate redundant computation. Additionally, systematic optimizations specifically designed for this workload and memory access pattern are applied to the MoE execution flow, further enhancing overall efficiency.
Evaluation results demonstrate that Samoyeds outperforms SOTA solutions in both computation speed and memory efficiency, while also providing superior model accuracy and hardware compatibility.

\section{Acknowledgments}
We thank all the anonymous reviewers and our shepherd, Dr. Qian Ge, for their insightful and detailed suggestions.
This work was funded by the National Key Research \& Development Program of China (No. 2022YFB4502002), the project from Wuxi Institute of Advanced Technology, NSFC (No. 62032008), STCSM (No. 23511100100), and Shanghai Science and Technology Development Funds (22QB1404600). This work is also supported by the Embedded Common Basic Software Technology Innovation Center. The corresponding authors are Heng Shi and Jianguo Yao.

\bibliographystyle{ACM-Reference-Format}
\bibliography{sample-base}

\newpage
\appendix
\section{Artifact Appendix} 
% \textit{This artifact appendix is meant to be a self-contained document which describes a roadmap for the evaluation of your artifact. It should include a clear description of the hardware, software, and configuration requirements as well as the major claims made by your paper and how to reproduce each claim through your artifact. Linking the claims of your paper to the artifact is a necessary step that ultimately allows artifact evaluators to reproduce your results. Towards that end, you should explicitly list down items (e.g., results, plots, tables) from the paper and cross-reference those with the experiments to be reproduced with your artifact.}\\
% \textit{Please fill all the mandatory sections, keeping their titles and organization but removing the current illustrative content, and remove the optional sections \ref{sec:reuse} and \ref{sec:gnotes} where those do not apply to your artifact.}

%%%%%%%%%%%%%%%%%%%%%%%%%%%%%%%%%%%%%%%%%%%%%%%%%%%%%%%%%%%%%%%%%%%%%
\subsection{Abstract}
% {\em [Mandatory]} 
% {\em Provide a short description of your artifact.}
This artifact includes the source codes and experiments for replicating the evaluations in this paper.

%%%%%%%%%%%%%%%%%%%%%%%%%%%%%%%%%%%%%%%%%%%%%%%%%%%%%%%%%%%%%%%%%%%%%
\subsection{Description \& Requirements}

% \textit{[Mandatory] This section should list all the information necessary to recreate the same experimental setup you have used to run your artifact. This includes at least a persistent link to a publicly accessible archival repository where all the artifact's main components (software, data-sets, documentation, etc.) can be accessed and, where this apply, the minimal hardware and software requirements to run your artifact. It is also very good practice to list and describe in this section benchmarks where those are part of, or simply have been used to produce results with, your artifact.}

\subsubsection{How to access}
% \textit{Describe here how to access your artifact. In case of a public repository, you should provide a persistent link to it. In case of a private repository, you should provide instructions on how to access it and where that access will be limited to the duration of this evaluation, that should be clearly indicated.\\}
% Note: This evaluation do not mandate the use of specific public repositories, so institutional repositories, or open commercial repositories are acceptable. In any case, repositories used to archive the artifact should have a declared plan to enable permanent accessibility.
All the source code and instructions can be accessed through the following platforms:
\begin{itemize}
    \item \textbf{git}: \url{https://github.com/guqiqi/Samoyeds.git}
    \item \textbf{zenodo}: \url{https://doi.org/10.5281/zenodo.14880516}
    \item \textbf{docker image}: kevinwu2017/samoyeds:1.0.0
\end{itemize}

\subsubsection{Hardware dependencies}
GPUs with Sparse Tensor Core (such as NVIDIA GPUs with Ampere architecture or newer).

\subsubsection{Software dependencies}
% We recommend running the Samoyeds artifact on Linux platform. The artifact is prepacked in a docker image, so the target platform should equipped with docker and NVIDIA GPU driver that supports CUDA 11.4+.

We recommend running the Samoyeds artifact on a Linux platform with Docker and an NVIDIA GPU driver supporting CUDA 11.4+. The artifact is pre-packaged in a Docker image.

\subsubsection{Benchmarks} 
% \textit{Describe here any data (e.g., data-sets, models, workloads, etc.) required by the experiments with this artifact reported in your paper.} \textit{[Simply write "None." where this does not apply to your artifact.]}
The Samoyeds artifact requires several models and datasets, such as Bert, SQuAD 1.1, etc. All of these requirements will be automatically downloaded during runtime.

%%%%%%%%%%%%%%%%%%%%%%%%%%%%%%%%%%%%%%%%%%%%%%%%%%%%%%%%%%%%%%%%%%%%%
\subsection{Set-up}

% {\em [Mandatory]} \textit{This section should include all the installation and configuration steps required to prepare the environment to be used for the evaluation of your artifact.}

\begin{lstlisting}[language=bash]
docker pull kevinwu2017/samoyeds:1.0.0
docker run -it --gpus all --name samoyeds-ae kevinwu2017/samoyeds:1.0.0
\end{lstlisting}

%%%%%%%%%%%%%%%%%%%%%%%%%%%%%%%%%%%%%%%%%%%%%%%%%%%%%%%%%%%%%%%%%%%%%
\subsection{Evaluation workflow}
% {\em [Mandatory]} \textit{This section should include all the operational steps and experiments which must be performed to evaluate your artifact is functional and to validate your paper's key results and claims. For that purpose, we ask you to use the two following subsections and cross-reference the items therein as explained next.}

\subsubsection{Major Claims}

\begin{itemize}
    \item \textbf{Claim (C1)}: Samoyeds achieves an average speedup of 1.99$\times$ over baselines, as shown in Figure 12 and 13. This is proven by experiment (E1).
    \item \textbf{Claim (C2)}: Samoyeds outperforms the baseline by 1.45$\times$ on average, as shown in Figure 14. This is validated by experiment (E2).
    \item \textbf{Claim (C3)}: Samoyeds 1.42$\times$ improves overall model performance by 1.42$\times$ (Figure 15) and delivers superior throughput across different batch sizes (Figure 16). This is confirmed by experiment (E3).
    % The overall model performance of Samoyeds is 1.42$\times$ better than baselines, as shown in Figure 15, and the throughput of Samoyeds under different batch size configurations outperforms other baselines, as shown in Figure 16. This is proven by experiment (E3).
    \item \textbf{Claim (C4)}: Different optimizations of Samoyeds provide speedup according to our breakdown analysis (Figure 17). This can be reproduced with experiment (E4).
    \item \textbf{Claim (C5)}: The optimization of Samoyeds does not affect the model accuracy, as shown in Table 4 and 5. This is verified by experiment (E5).
    \item \textbf{Claim (C6)}: Samoyeds exhibits superior portability compared to baselines, as shown in Figure 18. This is proven by experiment (E6).
    % The portability of Samoyeds surpasses baselines, as shown in Figure 18. This is proven by experiment (E6).
\end{itemize}

\subsubsection{Experiments}

The hardware requirements for each experiment are as follows:
\begin{itemize}
    \item E1, E2, E3, and E6: These experiments can be conducted on a single GPU, such as the NVIDIA GeForce RTX 4070 Super used in our paper.
    \item E4: This experiment involves post-training of models, which may require high-end GPUs such as the A100-80G used in our paper.
    \item E5: This experiment analyzes performance portability and requires multiple GPUs with different architectures (e.g., RTX 3090, RTX 4070 Super, RTX 4090, and A100, as used in our paper).
\end{itemize}

\begin{itemize}
\item \textbf{Experiment (E1)}: To reproduce the kernel level results (Figure 12, 13), execute:

\begin{lstlisting}[language=bash]
./artifacts/kernel/synthetic_scripts.sh
./artifacts/kernel/kernel_model_config_scripts.sh
\end{lstlisting}

Figure 12 and 13 can be plotted with following files:
\begin{lstlisting}[language=bash]
./artifacts/kernel/figure12_plot.ipynb
./artifacts/kernel/figure13_plot.ipynb
\end{lstlisting}

\item \textbf{Experiment (E2)}: To reproduce the MoE module level results (Figure 14), execute:

\begin{lstlisting}[language=bash]
./artifacts/MoE/figure14_scripts.sh
\end{lstlisting}

Figure 14 can be plotted with following files:
\begin{lstlisting}[language=bash]
./artifacts/MoE/figure14_plot.ipynb
\end{lstlisting}

\item \textbf{Experiment (E3)}: To reproduce the end-to-end level results (Figure 15, 16), execute:

\begin{lstlisting}[language=bash]
./artifacts/model/figure15_scripts.sh
./artifacts/model/figure16_scripts.sh
\end{lstlisting}

Figure 15 and 16 can be plotted with following files:
\begin{lstlisting}[language=bash]
./artifacts/model/figure15_plot.ipynb
./artifacts/model/figure16_plot.ipynb
\end{lstlisting}

\item \textbf{Experiment (E4)}: To reproduce the breakdown analysis results (Figure 17), execute:

\begin{lstlisting}[language=bash]
./artifacts/MoE/figure17_scripts.sh
\end{lstlisting}

Figure 17 can be plotted with following files:
\begin{lstlisting}[language=bash]
./artifacts/MoE/figure17_plot.ipynb
\end{lstlisting}

\item \textbf{Experiment (E5)}: We provide several scripts to reproduce the results of model accuracy (Table 4, 5). The following scripts require execution on high-memory GPUs or multi-GPU configurations. Specifically: (1) The script for collecting data in Table 4 is configured to utilize a cluster of 4 GPUs; (2) The scripts for collecting data in Table 5 must be run on an NVIDIA A100 80GB GPU to avoid Out-Of-Memory (OOM) errors. Lower-capacity GPUs may not have sufficient memory to handle these operations.

\begin{lstlisting}[language=bash]
cd sparseml
# Table 4
bach benchmark/scripts/samoyeds_gradual_pair.sh
# Table 5
bash benchmark/scripts/samoyeds_qwen2_80G.sh
bash benchmark/scripts/samoyeds_tiny_llama_80G.sh
\end{lstlisting}

The results are stored in the \textit{./benchmark/output\_dir/} directory.

\item \textbf{Experiment (E6)}: To reproduce the performance portability results of Samoyeds (Figure 18), the following script need to run on multiple GPUs, including NVIDIA GeForce RTX 3070, NVIDIA GeForce RTX 4070 Super, NVIDIA GeForce RTX 4090, and NVIDIA A100. 

\begin{lstlisting}[language=bash]
./artifacts/kernel/synthetic_scripts.sh
\end{lstlisting}

Figure 18 can be reproduced by collecting results on different GPUs into \textit{./artifacts/results/kernel/} folder.

Figure 18 then can be plotted with following files:
\begin{lstlisting}[language=bash]
./artifacts/MoE/figure18_plot.ipynb
\end{lstlisting}

\end{itemize}

\end{document}